\documentclass{article}

\usepackage{microtype}
\usepackage{graphicx}
\usepackage{subcaption}
\usepackage{booktabs} 

\usepackage[table]{xcolor} 
\usepackage{hyperref}

\usepackage[preprint]{icml2026}

\usepackage{amsmath}
\usepackage{amssymb}
\usepackage{mathtools}
\usepackage{amsthm}
\usepackage{multicol}
\usepackage{multirow}

\usepackage[capitalize,noabbrev]{cleveref}

\theoremstyle{plain}

\theoremstyle{definition}

\theoremstyle{remark}

\usepackage[textsize=tiny]{todonotes}

\icmltitlerunning{Chem4DLLM: 4D Multimodal LLMs for Chemical Dynamics Understanding}

\begin{document}

\twocolumn[
  \icmltitle{Chem4DLLM: 4D Multimodal LLMs for Chemical Dynamics Understanding}



  \icmlsetsymbol{equal}{*}

  \begin{icmlauthorlist}
    \icmlauthor{Xinyu Li}{equal,yyy}
    \icmlauthor{Zhen Zhang}{equal,yyy}
    \icmlauthor{Qi Chen }{yyy}
    \icmlauthor{Anton van den Hengel}{yyy}
    \icmlauthor{Lina Yao}{csiro}
    \icmlauthor{Javen Qinfeng Shi}{yyy}
  \end{icmlauthorlist}

  \icmlaffiliation{yyy}{Australian Institute for Machine Learning, Adelaide University, Adelaide, Australia}
  \icmlaffiliation{csiro}{CSIRO's Data61, Sydney, Australia}

  \icmlcorrespondingauthor{Xinyu Li}{henry.li@adelaide.edu.au}

  \icmlkeywords{Machine Learning, ICML}

  \vskip 0.3in
]



\printAffiliationsAndNotice{}  

\begin{abstract}

Existing chemical understanding tasks primarily rely on static molecular representations, limiting their ability to model inherently dynamic phenomena such as bond breaking or conformational changes, which are essential for a chemist to understand chemical reactions. To address this gap, we introduce \textbf{Chem}ical \textbf{D}ynamics \textbf{U}nderstanding (ChemDU), a new task that translates 4D molecular trajectories into interpretable natural-language explanations. ChemDU focuses on fundamental dynamic scenarios, including gas-phase and catalytic reactions, and requires models to reason about key events along molecular trajectories, such as bond formation and dissociation, and to generate coherent, mechanistically grounded narratives. 
To benchmark this capability, we construct \textbf{Chem4DBench}, the first dataset pairing 4D molecular trajectories with expert-authored explanations across these settings. We further propose \textbf{Chem4DLLM}, a unified model that integrates an equivariant graph encoder with a pretrained large language model to explicitly capture molecular geometry and rotational dynamics. We hope that ChemDU, together with Chem4DBench and Chem4DLLM, will stimulate further research in dynamic chemical understanding and multimodal scientific reasoning.

\end{abstract}

\begin{figure}
    \centering
    \includegraphics[width=0.95\linewidth]{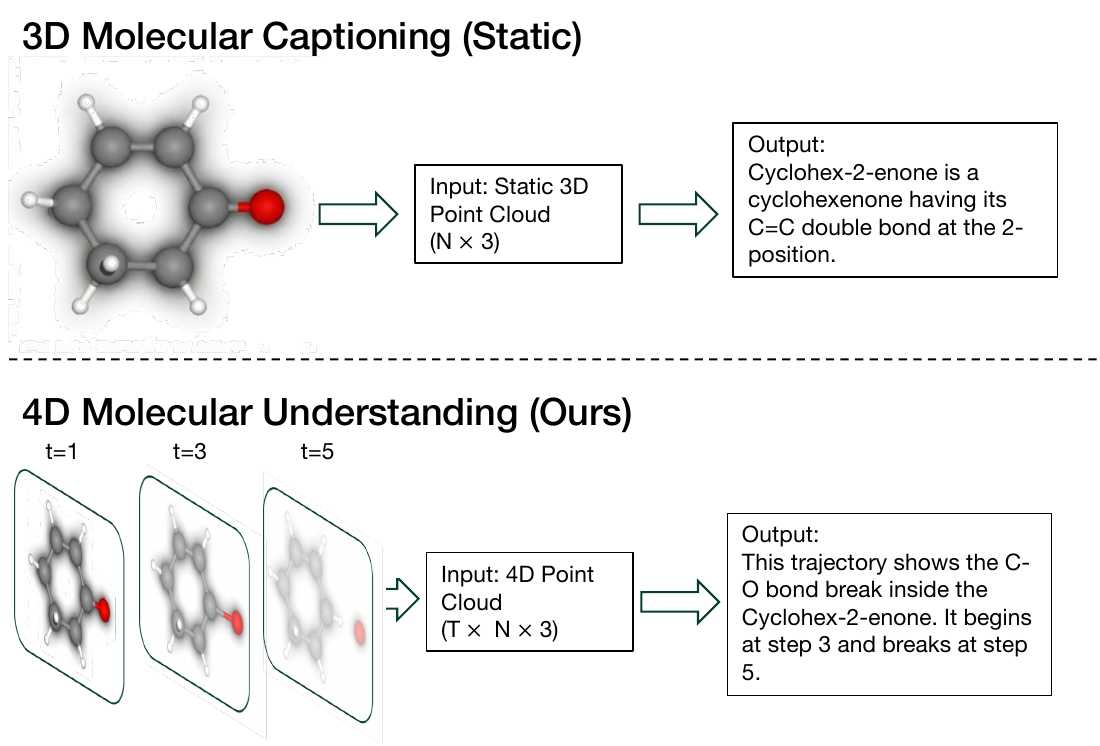}
    \caption{Comparison between static 3D molecular captioning and our proposed 4D molecular understanding task. (Top) Previous 3D methods utilize a static point cloud input $N \times 3$ to identify the molecule, such as Cyclohex-2-enone. (Bottom) Our 4D approach processes a temporal sequence of point clouds $T \times N \times 3$, enabling the model to describe dynamic chemical events. In this example, the model correctly identifies the trajectory of a C-O bond breaking within the Cyclohex-2-enone molecule, noting that the process initiates at $t=3$ and completes by $t=5$.}
    \label{fig:placeholder}
\end{figure}

\section{Introduction}
\label{sec:intro}
The advancement of large language models (LLMs)~\citep{achiam2023gpt,chen2024weak} has triggered a wave of innovation, moving these models from processors of human language to powerful reasoning engines for complex scientific data \citep{ai4science2023,agenticscience,Xin2025}. In chemistry and biology, this has sparked a new frontier of molecule-text modeling. Early efforts successfully unified textual knowledge with 1D molecular sequences (like SMILES \citep{smiles} and SELFIES \citep{selfies,chemt5} and/or 2D molecular graphs \citep{molfm} enabling powerful models for tasks like molecule captioning and text-based retrieval \citep{text2mol}. More recently, research has pushed into incorporating static 3D structures \citep{3dmolm,3dmolt5}, recognizing that a molecule's geometry is critical to its function. However, these static snapshots fail to capture the true nature of biomolecular processes, which are fundamentally dynamic. Indeed, all chemical phenomena are dynamic since an atomic system can be truly static only at absolute zero temperature, which does not exist in any known physical or biological environment.


Computational chemistry simulations such as Molecular Dynamics (MD) simulates such dynamic processes, generating massive spatio-temporal datasets that describe the fundamental mechanisms of biology and materials science. These trajectories, which can be conceptualized as dynamic 4D point clouds—where atomic coordinates (3D) evolve with time steps (1D)—hold the key to understanding processes like drug binding, protein folding, and catalytic reactions. Currently, such data is primarily interpreted through low-dimensional, quantitative metrics such as root-mean-square deviation or radial distribution functions. While useful, these metrics are insufficient; they fail to capture the qualitative, event-driven "story" of the molecular process. The true scientific insight lies in a high-level narrative: for example, "the inhibitor first docks into the active site, inducing a subtle conformational change in a nearby loop, which in turn allosterically closes the binding pocket." Automatically generating such qualitative descriptions is not only fundamental for human scientific insight but also crucial for agentic scientific discovery, where an agent is expected to plan, execute, and refine both in-silico simulations and in-vitro experiments. However, translating these vast, high-dimensional, and unstructured spatio-temporal data into coherent narratives presents a formidable challenge for the computer vision, machine learning, and computational chemistry communities.

The challenge comes from two sides. On the data side, while using AI and machine learning to accelerate quantum chemistry simulations is a community focus and has generated plentiful molecular, protein, and materials 3D geometries or 4D trajectories \citep{qm9,mp_2013,oc20,oc22}, corresponding text descriptions are rare. ChEBI-20 \citep{text2mol} and 3D-MoIT \citep{3dmolm} are two datasets that do contain 3D geometry and a text description, but both of them are focused on describing functional groups and static properties, therefore lacking descriptions of dynamic events. Another issue is that both datasets are designed for non-Periodic-Boundary-Condition (non-PBC) molecular systems, thereby lacking the capability to model PBC systems such as crystals and catalytic systems.

Beyond the data-side challenges, there are also unique challenges on the model side. First, the input is a high-dimensional, unstructured sequence of point clouds, how to encode such discrete graph data into LLM. Second, the sheer scale of the data is prohibitive for standard sequence models. A naive tokenization of continuous coordinates for hundreds of atoms over one hundred steps would generate sequences  that vastly exceed the context window of most of LLM. Third, 4D models are required to track state evolution, identify actions and events, and handle long-range dependencies—requirements significantly exceeding the static state description typical of 3D models. Finally, the target descriptions are not simple captions but requires understaning the causality, for example, one conformational changes is caused by an earlier binding event. 

To address these challenges, we make two concrete contributions. 
First, we introduce Chem4DBench, a benchmark designed to systematically evaluate large language models on 4D understanding. Chem4DBench focuses on a basic 4D scenerious -- reaction-- to assess whether a model can consume, condition on, and reason over molecular trajectories that include temporal evolution. 
The benchmark spans two controlled settings—gas-phase reactions and heterogeneous catalysis—each isolating different aspects of spatio-temporal reasoning. It covers both isolated molecular systems and periodic boundary condition (PBC) systems, while remaining grounded in physically meaningful simulation data.

Second, we propose Chem4DLLM, a baseline architecture that enables a pretrained LLM to directly condition on 4D molecular trajectories at atom-level resolution and generate structured textual descriptions. Chem4DLLM does not aim to replace quantum chemistry or molecular simulation pipelines. Instead, it serves as a diagnostic model for studying how current LLMs integrate geometric, temporal, and chemical information, and where their limitations lie when confronted with physically grounded 4D inputs.
Together, Chem4DBench and Chem4DLLM establish a realistic evaluation framework for progress toward interpretable and temporally grounded molecular reasoning, and aim to be the eye for LLM-driven agentic simulations.

\section{Related Work}
\subsection{Chemistry Large Language Models}\label{sec:chemllm}
The application of LLMs~\citep{achiam2023gpt,yang2025separation,ren2025efficiently} to chemistry necessitates a critical dimensional progression in molecular representation, moving from 1D sequential strings to 2D topological graphs, and further to 3D geometric structures. Initial efforts treated molecules as one-dimensional (1D) strings using notations like SMILES \citep{smiles} and SELFIES \citep{selfies}. These approaches adapted existing NLP architectures, often T5-based \citep{Raffel2020}, for sequence-to-sequence tasks such as property prediction and molecule-to-text/text-to-molecule generation \citep{text2mol,molt5}. Notable foundational models in this direction include MolT5 \citep{molt5}, ChemT5 \citep{chemt5}, BioT5 \citep{biot5}, $\text{BioT5}^+$ \citep{biot5p} and ChemLLM \citep{chemllm}, which were jointly trained on 1D molecular sequences and text. Furthermore, instruction tuning efforts, such as Mol-Instructions \citep{molinst} and LlaSMol \citep{llasmol}, have also been tried. However, 1D models suffer from fundamental limitations, including multi-injection, sequence ambiguity (a single molecule having multiple valid SMILES strings), and a high frequency of syntactically or chemically invalid sequences.

To address these issues, the field advanced to 2D graph-based LLMs, which integrate Graph Neural Networks (GNNs) or external graph encoding modules to explicitly infuse topological knowledge. Several works utilize cross-modal learning to align 2D molecular graphs with text, including MoleculeSTM \citep{molstm}, MolFM \citep{molfm}, MolCA \citep{molca}, and MolX \citep{molx}. UniMoT \citep{unimot} further proposes a Vector Quantization-driven tokenizer to convert 2D molecular graphs into tokens for unified modeling with text. ChemVLM \citep{chemvlm} leverages pretrained VLMs to understand 2D molecular structure by transferring the 2D graph into an image. More recent endeavors incorporate 3D molecular information to capture spatial arrangements. MolBind \citep{molbind} aligns a 2D graph encoder, a 3D structure encoder, and a language encoder via contrastive learning. 3D-MoLM \citep{3dmolm} equips an LLM with an external Uni-Mol encoder  \citep{unimol} for molecule-to-text interpretation. In contrast, 3D-MolT5 utilizes a classical 3D feature engineering method, E3FP \citep{e3fp}, to replace the neural network-based Uni-Mol encoder in 3D-MoLM. Except for using encoder or classical geometrical representations, Chem3DLLM \citep{chem3dllm} used a method SDF2Text which direct transfer the 3D coordinates into text, to make it compatible with LLM. One similar approach is CrystalLLM \citep{crystal_llm}, which directly output crystal CIF format text for crystal generation. 
Despite recent advancements, most existing frameworks remain tailored for static structural analysis and lack a unified mechanism to encode the temporal evolution in 4D molecular trajectories.

\subsection{Spatio-temporal Language Benchmarks}
The task of translating complex, continuous spatio-temporal data into natural language has been explored in several machine learning domains other than chemistry. The most direct analogue to molecular motion is the field of 3D human motion captioning. Foundational work, such as the KIT Motion-Language dataset \citep{kit_2016}, established the pairing of 3D human pose sequences with simple, descriptive sentences. More recently, large-scale benchmarks like HumanML3D \citep{humanml3d_2022} have emerged, leveraging vast motion capture repositories such as AMASS \citep{amass_2019}. These datasets provide a rich corpus of 3D human motions, typically represented as parametric body models or joint coordinates, paired with descriptive textual annotations. While invaluable, these benchmarks primarily focus on holistic, descriptive captions of a single agent's actions. In the broader context, video-language understanding is a mature field supported by large-scale benchmarks, including MSR-VTT \citep{msrvtt_2016} and VATEX \citep{Wang_2019_ICCV}. These datasets, however, pair 2D video clips with textual annotations. Consequently, they operate on 2D projections of the 3D world, lacking the precise, high-dimensional coordinate data inherent to our domain.


\subsection{Representation Learning for 4D and 3D Data}
Machine learning for molecular modeling has transitioned from relying on expert-designed features, such as Coulomb Matrices \citep{cm_2012}, toward the use of GNNs. This field progressed from invariant models, such as SchNet \citep{schnet_2017, schnet_2018} and DimeNet \citep{dimenet_2020}, eventually leading to equivariant models like NequIP \citep{nequip_2022} and MACE \citep{mace_2022}. These equivariant architectures, which respect the necessary physical symmetry of rotation, have achieved state-of-the-art results in predicting quantum-chemical properties from static 3D geometries. More recently, universal atomic models—including M3GNet \citep{m3gnet}, MACE-MP \citep{mace_mp}, and UMA \citep{uma}—have emerged. These models aim to be trained on massive, diverse datasets to provide a general-purpose potential that does not require task-specific fine-tuning. However, the primary goal of these GNNs is simulation rather than interpretation; they are trained to predict atomic forces to generate a trajectory, replacing computationally expensive quantum chemical calculations.

Make LLM understand 3D or 4D data requires 3D features or representations. Previous 3D-LLM works, such as 3D-MoLM \citep{3dmolm} and 3D-MolT5 \citep{3dmolt5}, primarily utilize invariant features. Specifically, 3D-MoLM employs the Uni-Mol encoder \citep{unimol} which uses invariant spatial positional encodings , while 3D-MolT5 adopts the E3FP fingerprinting algorithm \citep{e3fp}. These features are suitable for describing static 3D structures as they ensure the output is invariant to global translations and rotations. However, in 4D spatiotemporal scenarios, reliance on purely invariant features presents a significant limitation: they cannot distinguish between different rotational states of a molecule. To address this, equivariant GNNs—specifically those incorporating at least $l=1$ features—are necessary for 4D LLMs to comprehend molecular rotations and dynamics.

\begin{figure*}
    \centering
    \includegraphics[width=0.8\linewidth]{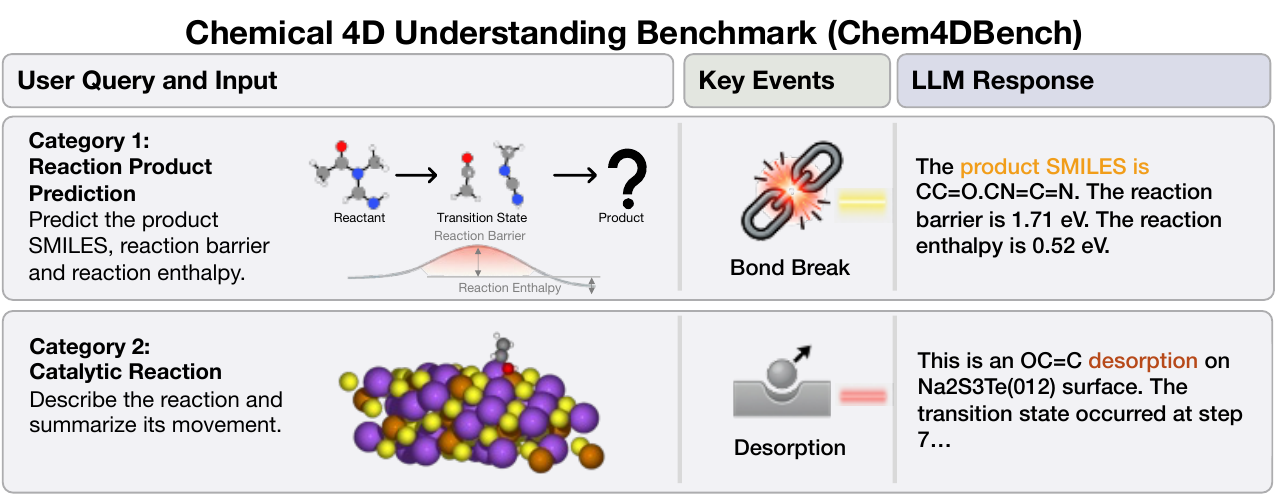}
    \caption{\small Overview of the Chem4D benchmark. This benchmark is a suit which encompasses three distinct categories: (1) Reaction Product Prediction, involving the analysis of bond breaking/forming events and reaction barriers (derived from Transition1x and RGD1); (2) Catalytic Reaction, covering complex surface interactions such as desorption (derived from OC20). For each category, the figure illustrates the workflow from the user query and 4D input, through the identification of key physical events, to the generation of the final scientific narrative.}
    \label{fig:placeholder}
\end{figure*}

\section{ChemDU Task and Chem4DBench}

\subsection{Problem Definition for ChemDU Task}

We define Chemical Dynamics Understanding (ChemDU) as the task of translating a 4D molecular trajectory into a concise, scientifically grounded natural-language description of the underlying chemical processes.

Formally, a trajectory is represented as
\[
\mathcal{T} = \{\mathcal{X}_t\}_{t=1}^{T},
\]
where each frame $\mathcal{X}_t = (\mathbf{A}, \mathbf{C}_t)$ consists of a fixed set of atoms $\mathbf{A}$ and their 3D Cartesian coordinates $\mathbf{C}_t \in \mathbb{R}^{N \times 3}$ at time step $t$. The ChemDU task learns a mapping
\begin{equation}
f: \mathcal{T}, \mathcal{P} \rightarrow \mathcal{Y},    
\end{equation}
where $\mathcal{P}$ is a sequences of tokens known as prompts that describes the goal of dynamic understanding (e.g. reaction path prediction), and $\mathcal{Y}$ is a natural-language explanation summarizing the key chemical dynamics observed over time.

Unlike static molecule captioning, ChemDU focuses on temporal evolution and event-level abstraction, requiring models to identify salient events (e.g., molecular motion, bond breaking and formation, adsorption or conformational changes), determine when they occur, and describe their mechanistic progression. The generated descriptions emphasize key events and their consequence, rather than static structural properties.

By explicitly targeting temporal grounding, event abstraction, and next frame prediction, ChemDU extends prior 1D--3D molecule--text tasks to dynamic molecular systems, enabling language-based interpretation of molecular dynamics simulations.

\subsection{Chem4DBench}

Chem4DBench is a benchmark for evaluating LLMs on 4D dynamic trajectories, focusing on reaction trajectories with explicit transition states, temporal evolution, and periodic boundary conditions. It consists of two complementary categories: gas-phase Reaction Products Prediction, and Catalytic Reactions—designed to probe different aspects of spatio-temporal and chemical reasoning.

\textbf{Reaction Products Prediction}
Chemical reactions represent a fundamentally 4D scenario, where atoms moving from their initial positions, through high-energy transition states, and finally ending with product geometries. Conventional reaction benchmarks such as USPTO-50K \citep{uspto_500k, uspto}, largely derived from the USPTO patent dataset \citep{uspto}, and treat reactions as a simple mapping between molecular graphs or SMILES strings. 
Therefore, they lack the essential dynamics and physics—such as steric hindrance, electronic reorganization, and the 3D spatial orientation of functional groups. Crucially, USPTO-based tasks completely omit the transition state, the most critical bottleneck of a chemical transformation. 
By focusing only on the "before and after" of a reaction, these models fail to account for the kinetic barriers and the specific geometric path required for a reaction to occur. To bridge this gap, we designed the Reaction Product Prediction to explicitly model the reaction coordinate.

This task requires the model to predict the product geometry given a 2 step input (geometries of the reactants and the TS). By incorporating the TS geometry, the model is forced to learn the actual pathways of atom migration and bond rearrangement rather than relying on statistical patterns in patent literature. Furthermore, Chem4D extends evaluation to intrinsic physical properties critical for experimental chemistry: the reaction barrier ($\Delta E^\ddagger$), which determines the reaction rate, and the reaction enthalpy ($\Delta H$), which represents the net heat change and determines thermodynamic feasibility.

This task is constructed using two complementary reaction datasets: Transition1x \citep{tran1x} and RGD1 \citep{rgd1}. Both datasets provide explicit reactant, TS, and product geometries, along with reaction barriers and enthalpies, making them uniquely suitable for evaluating physically grounded reaction understanding. Transition1x contains nudged elastic band (NEB)-generated trajectories for $\sim$10k organic reactions involving C, H, N, and O atoms. In contrast, RGD1 emphasizes chemical diversity and generalization. It contains over 170k distinct reactions enumerated via graphically defined elementary steps, covering molecules with up to 10 heavy atoms. To rigorously assess extrapolative reasoning, we construct multiple evaluation splits: an in-domian (ID) test set for Transition1x, an ID test set and three out-of-distribution (OOD) splits—OOD-Reactants, OOD-Products, and OOD-Both for RGD1. These OOD splits are determined via Scaffold splitting \citep{scaffold_split}. By evaluating on molecules unseen during training, this framework shifts reaction modeling from pattern matching toward a physically grounded 4D understanding of kinetics and thermodynamics.

\textbf{Catalytic Reactions}
A problem of previous 3D-based LLMs is that they are designed for molecules therefore can not be used for periodic boundary systems such as crystals and catalysts. Therefore, we also designed the catalytic reactions category, which aims to evaluate a model’s ability to understand complex 4D molecular dynamics in heterogeneous catalysis, where reactions occur at solid surfaces and involve strong coupling between adsorbates and the catalyst lattice. 

This subset is constructed based on the OC20-NEB dataset \citep{oc20_neb}, an extension of the Open Catalyst 2020 dataset \citep{oc20} that provides NEB trajectories for surface reactions. OC20-NEB explicitly models transition pathways for three key classes of elementary surface reactions—transfer, dissociation, and desorption—by resolving the atomic-scale evolution from reactant states, through TS, to product configurations. To overcome the limited scale of OC20-NEB, which contains only around 700 NEB pathways, we expand this subset with approximately 6,000 additional surface-reaction trajectories. This expansion is started by sampling catalyst–adsorbate reactant states from the OC20 dataset \citep{oc20}. Then we randomly assign a reaction from transfer, dissociaiton and desorption to each sample and re-simulate it. All expanded sampled are relaxed using NEB optimization in together with the pretrianed UMA model \citep{uma}, and energies are recorded along the reaction coordinate to identify TS, reaction barriers, and reaction enthalpies. This augmentation yields a large and diverse set of kinetically meaningful surface reaction pathways, substantially strengthening the Catalytic Reactions category and enabling rigorous evaluation of 4D understanding for heterogeneous catalytic systems.

\textbf{Evaluation}  We evaluate models on both Chem4DBench categories using metrics that jointly assess language quality, structural correctness, and physical fidelity (see Appendix~\ref{app:eval} for details). For gas-phase reactions, we ask the model to predict the final product SMILES, and we measure product SMILES accuracy as well as the regression error for the reaction barrier and enthalpy. For catalytic reactions, predicting the final product is less meaningful (e.g., transfer and desorption can yield the same product); therefore, we ask the model to identify the transition-state step, as well as the reaction barrier and enthalpy. We report reaction type accuracy, adsorbate and product SMILES accuracy, and energetic MAE. 

\section{Model Architecture}

\begin{figure}
    \centering
    \includegraphics[width=1.0\linewidth]{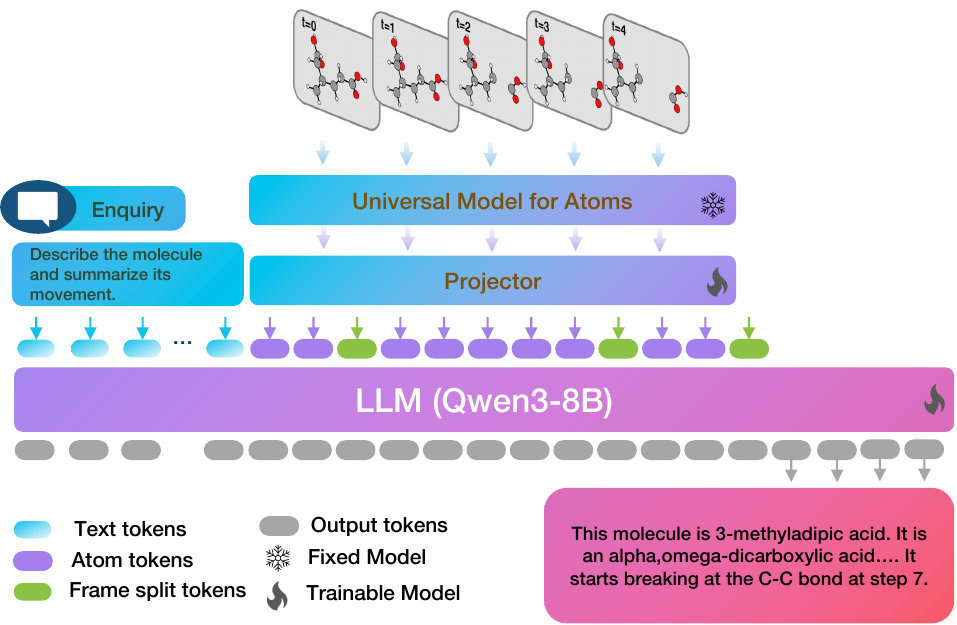}
    \caption{\small The Chem4DLLM model architecture.
(1) A 4D equivariant graph encoder (UMA) processes each 3D frame $\mathcal{X}_t$ into a graph embedding;
(2) A projector transforms the graph embeddings into vectors that are additively fused with the embeddings of the corresponding special \texttt{<graph>} tokens;
(3) The language model (Qwen3-8B) takes the resulting embedding sequence $\mathbf{E}$ as a prefix and autoregressively generates the output.}
    \label{fig:Chem4DLLM}
\end{figure}

Our model, built upon the Qwen3-8B architecture \citep{qwen}, is a multimodal LLM capable of processing and reasoning about 4D molecular trajectories. A key design choice is to use \emph{equivariant} representations (rather than rotationally invariant features): in 4D trajectories, rigid-body rotation and reorientation are meaningful dynamics, and invariant features discard this information, making distinct rotational states indistinguishable. By preserving how atomic features transform under 3D rotations, equivariant embeddings enable the model to track rotational dynamics and better localize geometry changes over time. Our architecture consists of three key components: (1) an equivariant graph encoder that translates each 3D frame into embeddings, (2) a graph projector, and (3) the Qwen3-8B LLM backbone.

\textbf{Equivariant Graph Encoder}
The input to our model is a 4D trajectory $\mathcal{T}$ represented as a sequence of atomic systems (frames). Each frame contains $N$ atoms. For each frame, we use UMA \citep{uma}, a pretrained equivariant graph neural network, to encode per-atom features. To retain rotational information while remaining computationally efficient, we use irreducible representations up to $l=1$, resulting in a 512-dimensional embedding per atom. Because EGNN-style message passing uses relative position differences (e.g., $\mathrm{pos}_i-\mathrm{pos}_j$), using only the UMA equivariant features would still be invariant to translations. Besides, these features primary describe the atomic environment and obscure the unit-cell information. To address this, we concatenate additional raw features, including atomic numbers, 3D Cartesian coordinates, PBC flags, and lattice cell vectors.

This yields a total feature dimensionality of 527 per atom. For a trajectory with $K$ frames, we represent the input as a sequence of node-embedding tensors $\mathbf{H} = [\mathbf{H}_1, \dots, \mathbf{H}_K]$, where each $\mathbf{H}_k \in \mathbb{R}^{N \times 527}$.

\textbf{Graph Projector}
We introduce three special tokens to mark the start of a graph, graph-node placeholders, and the end of a graph. For each trajectory frame, we serialize the graph as: \texttt{<graph\_start>}, followed by $N$ repeated \texttt{<graph>} placeholder tokens (one per atom), followed by \texttt{<graph\_end>}.

Within the model, for each \texttt{<graph>} placeholder token, the projected per-atom embeddings—obtained by mapping the 527-dimensional per-atom embeddings into the 4096-dimensional hidden space of Qwen3-8B via a learnable linear projector—are additively fused with the corresponding token embedding.
This design allows the LLM to attend directly to individual atoms and their features across multiple time steps through standard self-attention. In contrast to prior approaches that rely on temporal pooling or frame-level aggregation, our method preserves full atom-level resolution, enabling fine-grained reasoning about atomic interactions and their temporal evolution.

\paragraph{LLM Backbone}

The core of Chem4DLLM is the Qwen3-8B language model, implemented using a standard Transformer architecture with FlashAttention~2 for efficiency. 
The model processes a mixed sequence of text tokens and projected graph tokens. 
The full input sequence $Z$ is formed by concatenating the system prompt, the sequence of graph spans, and the user query or instruction. 
This formulation treats physical states as a foreign language that the LLM learns to interpret in context.

\paragraph{Training}

We perform full-parameter fine-tuning (or FSDP-based efficient fine-tuning) on the combined graph--text inputs. 
The model is trained using the standard causal language modeling (CLM) objective:
\begin{equation}
\mathcal{L} = -\sum_{t} \log p(y_t \mid y_{<t}, \mathbf{H}, \mathcal{P}),    
\end{equation}
where the node embedding tensors $\mathbf{H}$ is a proejcted representation of chemical trajectories $\mathcal{T}$, $\mathcal{P}$ is the prompt that describes the task, and  $y_t$ denotes the target text tokens (e.g., scientific descriptions or reasoning traces). 
Optimization is performed using AdamW with a cosine annealing learning rate schedule. 


Overall, Chem4DLLM treats a molecular trajectory as a temporally ordered sequence of physical states embedded directly into the token space of a causal language model. 
By preserving atom-level resolution across time and using autoregressive decoding, the model is forced to condition each generated token on all preceding structural states. 
This naturally induces a form of causal reasoning, where later textual descriptions must be grounded in earlier geometric and chemical events (e.g., bond formation, conformational rearrangements, or adsorption), enabling coherent narration of molecular processes rather than static captioning.


\section{Results}

\begin{table*}[t]
\centering
\scriptsize

\caption{Reaction product prediction results on Transition1x and RDG1. We evaluate models on the task of predicting product SMILES from reaction trajectories. BLEU measures sequence-level similarity between predicted and ground-truth SMILES. EXACT and Levenshtein distance assess string-level accuracy, while MACCS, RDK, and Morgan report fingerprint-based structural similarity. VALIDITY measures the fraction of chemically valid predictions, and MAE of reaction barriers and enthalpies evaluates energetic consistency. Results are reported for in-distribution (ID) and multiple out-of-distribution (OOD) settings with unseen reactants, products, or both. Overall, Chem4DLLM achieves strong performance across string, structural, and energetic metrics, particularly in challenging OOD regimes.}
\label{tab:t1x}
\begin{tabular}{lccccccccc}
Model           & BLEU$\uparrow$ & EXACT$\uparrow$ & LEVENSHTEIN$\downarrow$ & MACCS$\uparrow$ & RDK$\uparrow$  & MORGAN$\uparrow$ & VALIDITY$\uparrow$ & MAE Barrier$\downarrow$ & MAE Enthalpy$\downarrow$ \\ \hline
\rowcolor{gray!20}\multicolumn{10}{c}{\textit{Transition1x In Distribution}}                                                                                                                                                                \\ \hline
3D-MoLM & 0.216 & 0.075 & 10.385 & 0.296 & 0.258 & 0.168 & 0.514 & 3.157 & 3.618 \\
3D-MolT5 & 0.431 & 0.070 & 4.566 & 0.514 & 0.400 & 0.240 & 0.992 & 0.902 & 1.392 \\
Chem3D LLM & 0.059 & 0.000 & 8.347 & 0.206 & 0.124 & 0.053 & 0.954 & 1.262 & 2.064 \\
4D-MolT5 & 0.480 & 0.131 & 4.381 & 0.568 & 0.477 & 0.308 & 0.990 & 0.900 & 1.265 \\
4D Text-based & 0.051 & 0.000 & 8.318 & 0.200 & 0.119 & 0.049 & 0.957 & 1.552 & 2.204 \\
\textbf{Chem4DLLM} & \textbf{0.785} & \textbf{0.582} & \textbf{1.633} & \textbf{0.809} & \textbf{0.765} & \textbf{0.677} & \textbf{0.995} & \textbf{0.150} & \textbf{0.505} \\

\hline
\rowcolor{gray!20}\multicolumn{10}{c}{\textit{RDG1 In Distribution}}                                                                                                                            \\ \hline
3D-MoLM & 0.182 & 0.042 & 12.472 & 0.344 & 0.213 & 0.222 & 0.501 & 1.295 & 1.205 \\
3D-MolT5 & 0.640 & 0.280 & 4.416 & 0.722 & 0.476 & 0.484 & \textbf{0.996} & 0.687 & 0.639 \\
Chem3D LLM & 0.273 & 0.003 & 8.106 & 0.214 & 0.087 & 0.083 & 0.634 & 0.903 & 0.782 \\
4D-MolT5 & 0.706 & 0.405 & 3.760 & 0.770 & 0.566 & 0.572 & 0.995 & 0.443 & 0.532 \\
4D Text-based & 0.258 & 0.002 & 8.418 & 0.211 & 0.084 & 0.080 & 0.648 & 0.823 & 0.784 \\
\textbf{Chem4DLLM} & \textbf{0.785} & \textbf{0.527} & \textbf{3.176} & \textbf{0.860} & \textbf{0.726} & \textbf{0.733} & 0.993 & \textbf{0.211} & \textbf{0.393} \\

\hline
\rowcolor{gray!20}\multicolumn{10}{c}{\textit{RDG1 Out of Distribution Reactants}}                                                                                                                            \\ \hline
3D-MoLM & 0.162 & 0.012 & 14.018 & 0.274 & 0.124 & 0.134 & 0.481 & 3.400 & 3.668 \\
3D-MolT5 & 0.531 & 0.203 & \textbf{5.564} & 0.600 & 0.343 & 0.353 & \textbf{0.980} & 0.879 & 1.204 \\
Chem3D LLM & 0.281 & 0.003 & 9.592 & 0.333 & 0.125 & 0.119 & 0.856 & 1.205 & 1.080 \\
4D-MolT5 & \textbf{0.535} & \textbf{0.227} & 5.648 & \textbf{0.604} & \textbf{0.360} & \textbf{0.367} & 0.954 & 0.797 & 1.253 \\
4D Text-based & 0.275 & 0.003 & 9.742 & 0.328 & 0.123 & 0.116 & 0.857 & 0.998 & \textbf{1.042} \\
\textbf{Chem4DLLM} & 0.416 & 0.124 & 8.640 & 0.579 & 0.308 & 0.307 & 0.954 & \textbf{0.738} & 1.369 \\

\hline
\rowcolor{gray!20}\multicolumn{10}{c}{\textit{RDG1 Out of Distribution Products}}                                                                                                                            \\ \hline
3D-MoLM & 0.149 & 0.002 & 13.377 & 0.284 & 0.102 & 0.091 & 0.598 & 1.906 & 2.051 \\
3D-MolT5 & 0.346 & 0.019 & \textbf{7.274} & 0.483 & 0.180 & 0.169 & \textbf{0.984} & 0.906 & 1.254 \\
Chem3D LLM & 0.180 & 0.000 & 13.048 & 0.247 & 0.086 & 0.058 & 0.844 & 0.998 & 1.354 \\
4D-MolT5 & 0.366 & 0.055 & 7.413 & 0.489 & 0.204 & 0.199 & 0.954 & 0.761 & 1.231 \\
4D Text-based & 0.179 & 0.000 & 13.111 & 0.243 & 0.084 & 0.057 & 0.847 & 0.994 & 1.417 \\
\textbf{Chem4DLLM} & \textbf{0.433} & \textbf{0.083} & 9.240 & \textbf{0.562} & \textbf{0.306} & \textbf{0.295} & 0.948 & \textbf{0.459} & \textbf{1.117} \\

\hline
\rowcolor{gray!20}\multicolumn{10}{c}{\textit{RDG1 Out of Distribution Both}}                                                                                                                            \\ \hline
3D-MoLM & 0.132 & 0.000 & 14.419 & 0.218 & 0.086 & 0.082 & 0.437 & 4.384 & 4.692 \\
3D-MolT5 & 0.356 & 0.021 & \textbf{7.702} & 0.479 & 0.188 & 0.177 & \textbf{0.959} & 0.994 & 1.454 \\
Chem3D LLM & 0.214 & 0.000 & 12.545 & 0.267 & 0.099 & 0.060 & 0.915 & 1.132 & 1.433 \\
4D-MolT5 & 0.338 & \textbf{0.046} & 8.200 & 0.460 & 0.198 & 0.193 & 0.860 & 1.004 & 1.560 \\
4D Text-based & 0.209 & 0.000 & 12.653 & 0.260 & 0.095 & 0.058 & 0.903 & 1.068 & 1.450 \\
\textbf{Chem4DLLM} & \textbf{0.418} & 0.046 & 10.092 & \textbf{0.520} & \textbf{0.249} & \textbf{0.234} & 0.903 & \textbf{0.710} & \textbf{1.379} \\

\hline
\end{tabular}
\end{table*}

Both of the proposed Chem4DBench and Chem4DLLM are designed for general chemical scenarios, covering isolated molecules and periodic catalysts. This setting extends beyond most existing 3D molecule-language models, which are typically limited to molecules, and makes many prior baselines inapplicable to all Chem4D tasks. We therefore selectively adapt suitable baselines for each task category, as detailed below.

\subsection{Baselines}
We compare our Chem4DLLM against several representative baselines that span both static 3D molecular understanding and na\"ive extensions to 4D trajectories. Specifically, we include three strong 3D molecular LLMs, 3D-MoLM \citep{3dmolm}, 3D-MolT5 \citep{3dmolt5}, and Chem3D-LLM \citep{chem3dllm}, as well as their na\"ive 4D variants.

\textbf{3D LLM Baselines.}
3D-MoLM and 3D-MolT5 are designed for static molecular representations and are therefore evaluated only on the Reaction Product Prediction task. Chem3D-LLM, although originally proposed for molecular understanding, encodes atomic positions in a textual form, which allows it to be naturally extended to periodic boundary condition (PBC) systems. We thus adapt Chem3D-LLM (details in Appendix~\ref{appen:model}) and evaluate it on both Reaction Product Prediction and Catalytic Reaction Understanding.

\textbf{4D LLM Baselines.}
We construct 4D-MolT5 by extending 3D-MolT5 \citep{3dmolt5} to the temporal setting, in which a molecular trajectory is serialized as a temporally ordered sequence of 3D molecular frames. This formulation leverages the pretrained 3D-MolT5 backbone, allowing the model to jointly attend to spatial structures and their temporal evolution without modifying the underlying architecture.
However, because 3D-MolT5 relies on E3FP-based rotationally invariant molecular representations, this baseline is fundamentally limited in modeling rigid-body molecular rotations across time steps and cannot handle periodic boundary condition (PBC) systems such as catalytic surfaces. We further include a 4D Text-based LLM baseline adapted from Chem3D-LLM \citep{chem3dllm}, which represents molecular trajectories purely as serialized textual descriptions of atomic coordinates and time indices. While conceptually simple, this representation suffers from severe input-length limitations; for instance, a trajectory with 10 frames and 200 atoms can easily exceed 20{,}000 tokens. To partially alleviate this issue, we apply a heuristic compression strategy that records atomic positions only when coordinate changes exceed a threshold of 0.1. Additional implementation details are provided in Appendix~\ref{appen:model}.

\subsection{Chemical Dynamics Understanding}
\begin{table*}[t]
\centering
\caption{\small Evaluation results on the catalytic reaction subset. Metrics cover reaction type classification (Type Acc), inital adsorbate and reaction product SMILES prediction using exact match and Morgan fingerprint similarity, transition-state timestep prediction (TS Step MAE), and MAEs for the predicted reaction barriers and enthalpies. Chem4DLLM consistently outperforms both 4D text-based method and Chem3DLLM baselines across both structural and energetic metrics, highlighting the importance of explicit spatio-temporal modeling for reaction dynamics.}

\label{tab:oc20_results}
\scriptsize
\begin{tabular}{lccccccccc}
Model      & Type Acc $\uparrow$ & Ads Exact $\uparrow$ & Ads Morgan $\uparrow$ & Prod Exact $\uparrow$ & Prod Morgan $\uparrow$ & TS Step MAE $\downarrow$ & Barrier MAE $\downarrow$ & Enthalpy MAE $\downarrow$ \\ \hline
Chem3D & 0.517                   & 0.355                & 0.392                 & 0.054                 & 0.064                                & 2.437                    & 0.985           & 1.069            \\
4D text-based & 0.535          & 0.373       & 0.410        & 0.068        & 0.084             & 1.947           & 1.019                    & 1.102                     \\ 
Chem4DLLM & \textbf{0.774}          & \textbf{0.762}       & \textbf{0.776}        & \textbf{0.432}        & \textbf{0.454}             & \textbf{1.348}           & \textbf{0.848}                    & \textbf{1.019}                     \\ \hline
\end{tabular}
\end{table*}

\textbf{Reaction Products Prediction} We begin our discussion by evaluating 4D models on the task of reaction product prediction and compare its performance against baselines mentioned above. As shown in Table~\ref{tab:t1x}, Chem4DLLM demonstrates a clear advantage in capturing the complex atomistic rearrangements of chemical reactions. On the Transition1x dataset, Chem4DLLM achieves a BLEU score of 0.785 and an EXACT match rate of 0.582, significantly outperforming the next-best, 4D-MolT5 (BLEU 0.480, EXACT 0.131). This superior performance extends to chemical structural metrics, where Chem4DLLM reaches a Morgan \citep{Morgan1965} fingerprint similarity of 0.677, nearly doubling the performance of 4D-MolT5 (0.308). In the larger and more chemically diverse RGD1 dataset, Chem4DLLM maintains its lead with an EXACT match rate of 0.527 in the ID split. Even in challenging OOD scenarios—such as when both reactants and products involve unseen molecular scaffolds—Chem4DLLM achieves a BLEU score of 0.418, which remains superior to the 0.338 achieved by 4D-MolT5.

Beyond structural prediction, Chem4DLLM excels at estimating intrinsic physical properties, which is crucial for downstream chemical applications. It achieves a MAE for the reaction barrier of only 0.150 eV on Transition1x, compared to 0.900 eV for 4D-MolT5. Similarly, the MAE for enthalpy is reduced to 0.505 eV, demonstrating that the model’s spatio-temporal event tokenizer effectively captures the energetic landscape of the reaction. 

The strong gains of Chem4DLLM suggest that a 4D formulation provides critical advantages for reaction modeling. Chemical reactions are inherently dynamical: bond breaking and formation arise from continuous atomic motion along a reaction coordinate, rather than from any single static structure. Static 3D models such as 3D-MolM \cite{3dmolm} and 3D-MolT5 \cite{3dmolt5} operate on isolated geometries and collapse this process into a snapshot, limiting their ability to distinguish inert structures from actively transforming ones.

In contrast, 4D models represent molecules as time-ordered trajectories, explicitly capturing the evolution from reactants to transition states. This temporal context is essential for differentiating reactions with similar initial geometries but distinct kinetic pathways, and explains the poor performance of 3D models on trajectory-dependent quantities such as reaction barriers and enthalpies. Because these properties are determined by the highest-energy configuration along the path, the 4D representation enables Chem4DLLM to encode energetic evolution directly, resulting in substantially lower prediction errors and a more physically grounded understanding of chemical reactivity.

\textbf{Catalytic Reaction Understanding} We further evaluate Chem4DLLM on catalytic reaction understanding using the OC20Bench, which represents a substantially more challenging setting involving PBC, surface–adsorbate interactions, and heterogeneous reaction pathways. As summarized in Table~\ref{tab:oc20_results}, Chem4DLLM consistently outperforms both static 3D and text-based 4D baselines across all evaluated metrics. In particular, Chem4DLLM achieves a reaction type accuracy of 0.774, significantly exceeding Chem3DLLM (0.517) and the 4D text-based method (0.535), indicating a stronger capability to recognize global catalytic reaction patterns. For adsorption structure prediction, Chem4DLLM attains an Adsorbate Exact match of 0.762 and a Morgan similarity of 0.776, demonstrating precise modeling of surface binding motifs and adsorbate conformations that are critical for catalytic reactivity.

Chem4DLLM also shows substantial gains in product SMILES prediction, reaching a Product Exact match rate of 0.432 and a Morgan similarity of 0.454, whereas all baselines struggle in this regime. This gap highlights the difficulty of inferring surface-mediated bond formation and breaking from static or weakly structured representations. By contrast, Chem4DLLM explicitly captures adsorption, diffusion, and reaction events along the trajectory, enabling more accurate reasoning over catalytic reaction pathways.

In addition to structural accuracy, Chem4DLLM exhibits improved performance on physically meaningful regression targets. It reduces the MAE for transition-state step prediction to 1.348 and achieves lower barrier and enthalpy MAEs compared to all baselines. These results suggest that the use of pretrained foundation model preserve essential energetic information and reflect the underlying reaction coordinate, rather than relying on superficial correlations.

Overall, the OC20 results demonstrate that Chem4DLLM generalizes beyond molecular-scale reactions to complex catalytic systems under PBC.

\section{Conclusion}
We introduced Chemical Dynamics understanding task, aiming to translate spatio-temporal molecular simulation data into coherent, scientifically grounded narratives. To support systematic evaluation, we presented Chem4DBench, the first benchmark that pairs 4D trajectories with expert-authored text for both gas-phase reactions and catalytic reactions.

We further proposed Chem4DLLM, a multimodal 4D-LLM framework that injects equivariant graph representations of atomic systems into an LLM via a lightweight projection interface. Across both gas-phase and catalytic reactions, Chem4DLLM demonstrates strong performance compared with both 3D baselines and their 4D variants, particularly on metrics that require temporal reasoning. Our results suggest that explicitly modeling 4D structure, rather than relying on static snapshots or long coordinate serializations, is critical for undersatnd dynamic process.

Looking forward, we believe Chem4DBench can serve as a foundation for future research on scalable 4D tokenization, long-horizon trajectory reasoning, and richer supervision beyond end-to-end narration (e.g., intermediate event discovery and mechanistic rationales). Extending coverage to more diverse chemistries, longer trajectories, and higher-fidelity simulation/experiment pairings is an important next step toward robust agentic scientific discovery.

\section*{Impact Statement}

This work introduces Chemical Dynamics Understanding benchmark , providing a framework to translate 4D molecular trajectories into interpretable scientific narratives. By automating the explanation of complex events like bond breaking , reaction pathways , and catalytic interactions , we expect to accelerate discovery in drug design and sustainable materials. This advancement supports agentic scientific discovery by helping models plan and refine simulations through high-level qualitative insights. While dual-use risks exist in molecular modeling, our goal is to enhance transparency and reasoning in Machine Learning for the broader benefit of the scientific community

\nocite{langley00}

\bibliography{example_paper}

@String(CVPR= {IEEE Conf. Comput. Vis. Pattern Recog.})

@String(ICCV= {Int. Conf. Comput. Vis.})

@String(ICLR = {Int. Conf. Learn. Represent.})

@String(AAAI = {AAAI})

@String(CVPR  = {CVPR})

@String(ICCV  = {ICCV})

@String(ICLR  = {ICLR})

@article{kit_2016,
    author = {Matthias Plappert and Christian Mandery and Tamim Asfour},
    title = {The {KIT} Motion-Language Dataset},
    journal = {Big Data},
    publisher = {Mary Ann Liebert Inc},
    year = 2016,
    month = {dec},
    volume = {4},
    number = {4},
    pages = {236--252},
    url = {http://dx.doi.org/10.1089/big.2016.0028},
    doi = {10.1089/big.2016.0028},
}

@InProceedings{humanml3d_2022,
    author    = {Guo, Chuan and Zou, Shihao and Zuo, Xinxin and Wang, Sen and Ji, Wei and Li, Xingyu and Cheng, Li},
    title     = {Generating Diverse and Natural 3D Human Motions From Text},
    booktitle = {Proceedings of the IEEE/CVF Conference on Computer Vision and Pattern Recognition (CVPR)},
    month     = {June},
    year      = {2022},
    pages     = {5152-5161}
}

@conference{amass_2019,
  title = {{AMASS}: Archive of Motion Capture as Surface Shapes},
  author = {Mahmood, Naureen and Ghorbani, Nima and Troje, Nikolaus F. and Pons-Moll, Gerard and Black, Michael J.},
  booktitle = {International Conference on Computer Vision},
  pages = {5442--5451},
  month = oct,
  year = {2019},
  month_numeric = {10}
}

@InProceedings{Wang_2019_ICCV,
author = {Wang, Xin and Wu, Jiawei and Chen, Junkun and Li, Lei and Wang, Yuan-Fang and Wang, William Yang},
title = {VaTeX: A Large-Scale, High-Quality Multilingual Dataset for Video-and-Language Research},
booktitle = {The IEEE International Conference on Computer Vision (ICCV)},
month = {October},
year = {2019}
}

@inproceedings{msrvtt_2016,
  author       = {Jun Xu and
                  Tao Mei and
                  Ting Yao and
                  Yong Rui},
  title        = {{MSR-VTT:} {A} Large Video Description Dataset for Bridging Video
                  and Language},
  booktitle    = {2016 {IEEE} Conference on Computer Vision and Pattern Recognition,
                  {CVPR} 2016, Las Vegas, NV, USA, June 27-30, 2016},
  pages        = {5288--5296},
  publisher    = {{IEEE} Computer Society},
  year         = {2016},
  url          = {https://doi.org/10.1109/CVPR.2016.571},
  doi          = {10.1109/CVPR.2016.571},
  bibsource    = {dblp computer science bibliography, https://dblp.org}
}

@article{smiles,
  title = {SMILES,  a chemical language and information system. 1. Introduction to methodology and encoding rules},
  volume = {28},
  ISSN = {1520-5142},
  url = {http://dx.doi.org/10.1021/ci00057a005},
  DOI = {10.1021/ci00057a005},
  number = {1},
  journal = {Journal of Chemical Information and Computer Sciences},
  publisher = {American Chemical Society (ACS)},
  author = {Weininger,  David},
  year = {1988},
  month = feb,
  pages = {31–36}
}

@article{selfies,
title = {SELFIES and the future of molecular string representations},
journal = {Patterns},
volume = {3},
number = {10},
pages = {100588},
year = {2022},
issn = {2666-3899},
doi = {https://doi.org/10.1016/j.patter.2022.100588},
url = {https://www.sciencedirect.com/science/article/pii/S2666389922002069},
author = {Mario Krenn and Qianxiang Ai and Senja Barthel and Nessa Carson and Angelo Frei and Nathan C. Frey and Pascal Friederich and Théophile Gaudin and Alberto Alexander Gayle and Kevin Maik Jablonka and Rafael F. Lameiro and Dominik Lemm and Alston Lo and Seyed Mohamad Moosavi and José Manuel Nápoles-Duarte and AkshatKumar Nigam and Robert Pollice and Kohulan Rajan and Ulrich Schatzschneider and Philippe Schwaller and Marta Skreta and Berend Smit and Felix Strieth-Kalthoff and Chong Sun and Gary Tom and Guido {Falk von Rudorff} and Andrew Wang and Andrew D. White and Adamo Young and Rose Yu and Alán Aspuru-Guzik},
}

@inproceedings{text2mol,
  title = {Text2Mol: Cross-Modal Molecule Retrieval with Natural Language Queries},
  url = {http://dx.doi.org/10.18653/v1/2021.emnlp-main.47},
  DOI = {10.18653/v1/2021.emnlp-main.47},
  booktitle = {Proceedings of the 2021 Conference on Empirical Methods in Natural Language Processing},
  publisher = {Association for Computational Linguistics},
  author = {Edwards,  Carl and Zhai,  ChengXiang and Ji,  Heng},
  year = {2021}
}

@inproceedings{molt5,
  author       = {Carl Edwards and
                  Tuan Manh Lai and
                  Kevin Ros and
                  Garrett Honke and
                  Kyunghyun Cho and
                  Heng Ji},
  editor       = {Yoav Goldberg and
                  Zornitsa Kozareva and
                  Yue Zhang},
  title        = {Translation between Molecules and Natural Language},
  booktitle    = {Proceedings of the 2022 Conference on Empirical Methods in Natural
                  Language Processing, {EMNLP} 2022, Abu Dhabi, United Arab Emirates,
                  December 7-11, 2022},
  pages        = {375--413},
  publisher    = {Association for Computational Linguistics},
  year         = {2022},
  url          = {https://doi.org/10.18653/v1/2022.emnlp-main.26},
  doi          = {10.18653/V1/2022.EMNLP-MAIN.26},
  bibsource    = {dblp computer science bibliography, https://dblp.org}
}

@inproceedings{chemt5,
  author       = {Dimitrios Christofidellis and
                  Giorgio Giannone and
                  Jannis Born and
                  Ole Winther and
                  Teodoro Laino and
                  Matteo Manica},
  editor       = {Andreas Krause and
                  Emma Brunskill and
                  Kyunghyun Cho and
                  Barbara Engelhardt and
                  Sivan Sabato and
                  Jonathan Scarlett},
  title        = {Unifying Molecular and Textual Representations via Multi-task Language
                  Modelling},
  booktitle    = {International Conference on Machine Learning, {ICML} 2023, 23-29 July
                  2023, Honolulu, Hawaii, {USA}},
  series       = {Proceedings of Machine Learning Research},
  volume       = {202},
  pages        = {6140--6157},
  publisher    = {{PMLR}},
  year         = {2023},
  url          = {https://proceedings.mlr.press/v202/christofidellis23a.html},
  bibsource    = {dblp computer science bibliography, https://dblp.org}
}

@inproceedings{biot5,
title={BioT5: Enriching Cross-modal Integration in Biology with Chemical Knowledge and Natural Language Associations},
author={Qizhi Pei and Wei Zhang and Jinhua Zhu and Kehan Wu and Kaiyuan Gao and Lijun Wu and Yingce Xia and Rui Yan},
booktitle={The 2023 Conference on Empirical Methods in Natural Language Processing},
year={2023},
url={https://openreview.net/forum?id=uhVJ3SLq80}
}

@inproceedings{biot5p,
  author       = {Qizhi Pei and
                  Lijun Wu and
                  Kaiyuan Gao and
                  Xiaozhuan Liang and
                  Yin Fang and
                  Jinhua Zhu and
                  Shufang Xie and
                  Tao Qin and
                  Rui Yan},
  editor       = {Lun{-}Wei Ku and
                  Andre Martins and
                  Vivek Srikumar},
  title        = {BioT5+: Towards Generalized Biological Understanding with {IUPAC}
                  Integration and Multi-task Tuning},
  booktitle    = {Findings of the Association for Computational Linguistics, {ACL} 2024,
                  Bangkok, Thailand and virtual meeting, August 11-16, 2024},
  pages        = {1216--1240},
  publisher    = {Association for Computational Linguistics},
  year         = {2024},
  url          = {https://doi.org/10.18653/v1/2024.findings-acl.71},
  doi          = {10.18653/V1/2024.FINDINGS-ACL.71},
  bibsource    = {dblp computer science bibliography, https://dblp.org}
}

@inproceedings{molca,
    title = "{M}ol{CA}: Molecular Graph-Language Modeling with Cross-Modal Projector and Uni-Modal Adapter",
    author = "Liu, Zhiyuan  and
      Li, Sihang  and
      Luo, Yanchen  and
      Fei, Hao  and
      Cao, Yixin  and
      Kawaguchi, Kenji  and
      Wang, Xiang  and
      Chua, Tat-Seng",
    editor = "Bouamor, Houda  and
      Pino, Juan  and
      Bali, Kalika",
    booktitle = "Proceedings of the 2023 Conference on Empirical Methods in Natural Language Processing",
    month = dec,
    year = "2023",
    address = "Singapore",
    publisher = "Association for Computational Linguistics",
    url = "https://aclanthology.org/2023.emnlp-main.966/",
    doi = "10.18653/v1/2023.emnlp-main.966",
    pages = "15623--15638",
}

@inproceedings{molinst,
  author       = {Yin Fang and
                  Xiaozhuan Liang and
                  Ningyu Zhang and
                  Kangwei Liu and
                  Rui Huang and
                  Zhuo Chen and
                  Xiaohui Fan and
                  Huajun Chen},
  title        = {Mol-Instructions: {A} Large-Scale Biomolecular Instruction Dataset
                  for Large Language Models},
  booktitle    = {The Twelfth International Conference on Learning Representations,
                  {ICLR} 2024, Vienna, Austria, May 7-11, 2024},
  publisher    = {OpenReview.net},
  year         = {2024},
  url          = {https://openreview.net/forum?id=Tlsdsb6l9n},
  bibsource    = {dblp computer science bibliography, https://dblp.org}
}

@inproceedings{llasmol,
title={Lla{SM}ol: Advancing Large Language Models for Chemistry with a Large-Scale, Comprehensive, High-Quality Instruction Tuning Dataset},
author={Botao Yu and Frazier N. Baker and Ziqi Chen and Xia Ning and Huan Sun},
booktitle={First Conference on Language Modeling},
year={2024},
url={https://openreview.net/forum?id=lY6XTF9tPv}
}

@article{molstm,
  title = {Multi-modal molecule structure–text model for text-based retrieval and editing},
  volume = {5},
  ISSN = {2522-5839},
  url = {http://dx.doi.org/10.1038/s42256-023-00759-6},
  DOI = {10.1038/s42256-023-00759-6},
  number = {12},
  journal = {Nature Machine Intelligence},
  publisher = {Springer Science and Business Media LLC},
  author = {Liu,  Shengchao and Nie,  Weili and Wang,  Chengpeng and Lu,  Jiarui and Qiao,  Zhuoran and Liu,  Ling and Tang,  Jian and Xiao,  Chaowei and Anandkumar,  Animashree},
  year = {2023},
  month = dec,
  pages = {1447–1457}
}

@misc{molfm,
      title={MolFM: A Multimodal Molecular Foundation Model}, 
      author={Yizhen Luo and Kai Yang and Massimo Hong and Xing Yi Liu and Zaiqing Nie},
      year={2023},
      eprint={2307.09484},
      archivePrefix={arXiv},
      primaryClass={q-bio.BM},
      url={https://arxiv.org/abs/2307.09484}, 
}

@misc{molx,
      title={MolX: Enhancing Large Language Models for Molecular Understanding With A Multi-Modal Extension}, 
      author={Khiem Le and Zhichun Guo and Kaiwen Dong and Xiaobao Huang and Bozhao Nan and Roshni Iyer and Xiangliang Zhang and Olaf Wiest and Wei Wang and Ting Hua and Nitesh V. Chawla},
      year={2025},
      eprint={2406.06777},
      archivePrefix={arXiv},
      primaryClass={cs.CV},
      url={https://arxiv.org/abs/2406.06777}, 
}

@misc{unimot,
      title={UniMoT: Unified Molecule-Text Language Model with Discrete Token Representation}, 
      author={Shuhan Guo and Yatao Bian and Ruibing Wang and Nan Yin and Zhen Wang and Quanming Yao},
      year={2025},
      eprint={2408.00863},
      archivePrefix={arXiv},
      primaryClass={cs.CL},
      url={https://arxiv.org/abs/2408.00863}, 
}

@inproceedings{3dmolm,
  author={Sihang Li and Zhiyuan Liu and Yanchen Luo and Xiang Wang and Xiangnan He and Kenji Kawaguchi and Tat-Seng Chua and Qi Tian},
  title={Towards 3D Molecule-Text Interpretation in Language Models},
  year={2024},
  cdate={1704067200000},
  url={https://openreview.net/forum?id=xI4yNlkaqh},
  booktitle={ICLR},
}

@misc{molbind,
      title={MolBind: Multimodal Alignment of Language, Molecules, and Proteins}, 
      author={Teng Xiao and Chao Cui and Huaisheng Zhu and Vasant G. Honavar},
      year={2024},
      eprint={2403.08167},
      archivePrefix={arXiv},
      primaryClass={cs.LG},
      url={https://arxiv.org/abs/2403.08167}, 
}

@inproceedings{
unimol,
title={Uni-Mol: A Universal 3D Molecular Representation Learning Framework},
author={Gengmo Zhou and Zhifeng Gao and Qiankun Ding and Hang Zheng and Hongteng Xu and Zhewei Wei and Linfeng Zhang and Guolin Ke},
booktitle={The Eleventh International Conference on Learning Representations },
year={2023},
url={https://openreview.net/forum?id=6K2RM6wVqKu}
}

@misc{3dmolt5,
      title={3D-MolT5: Leveraging Discrete Structural Information for Molecule-Text Modeling}, 
      author={Qizhi Pei and Rui Yan and Kaiyuan Gao and Jinhua Zhu and Lijun Wu},
      year={2025},
      eprint={2406.05797},
      archivePrefix={arXiv},
      primaryClass={q-bio.BM},
      url={https://arxiv.org/abs/2406.05797}, 
}

@article{Raffel2020,
  author  = {Colin Raffel and Noam Shazeer and Adam Roberts and Katherine Lee and Sharan Narang and Michael Matena and Yanqi Zhou and Wei Li and Peter J. Liu},
  title   = {Exploring the Limits of Transfer Learning with a Unified Text-to-Text Transformer},
  journal = {Journal of Machine Learning Research},
  year    = {2020},
  volume  = {21},
  number  = {140},
  pages   = {1--67},
  url     = {http://jmlr.org/papers/v21/20-074.html}
}

@misc{ai4science2023,
      title={The Impact of Large Language Models on Scientific Discovery: a Preliminary Study using GPT-4}, 
      author={Microsoft Research AI4Science and Microsoft Azure Quantum},
      year={2023},
      eprint={2311.07361},
      archivePrefix={arXiv},
      primaryClass={cs.CL},
      url={https://arxiv.org/abs/2311.07361}, 
}

@misc{agenticscience,
      title={From AI for Science to Agentic Science: A Survey on Autonomous Scientific Discovery}, 
      author={Jiaqi Wei and Yuejin Yang and Xiang Zhang and Yuhan Chen and Xiang Zhuang and Zhangyang Gao and Dongzhan Zhou and Guangshuai Wang and Zhiqiang Gao and Juntai Cao and Zijie Qiu and Ming Hu and Chenglong Ma and Shixiang Tang and Junjun He and Chunfeng Song and Xuming He and Qiang Zhang and Chenyu You and Shuangjia Zheng and Ning Ding and Wanli Ouyang and Nanqing Dong and Yu Cheng and Siqi Sun and Lei Bai and Bowen Zhou},
      year={2025},
      eprint={2508.14111},
      archivePrefix={arXiv},
      primaryClass={cs.LG},
      url={https://arxiv.org/abs/2508.14111}, 
}

@article{Xin2025,
  title = {Towards agentic science for advancing scientific discovery},
  volume = {7},
  ISSN = {2522-5839},
  url = {http://dx.doi.org/10.1038/s42256-025-01110-x},
  DOI = {10.1038/s42256-025-01110-x},
  number = {9},
  journal = {Nature Machine Intelligence},
  publisher = {Springer Science and Business Media LLC},
  author = {Xin,  Hongliang and Kitchin,  John R. and Kulik,  Heather J.},
  year = {2025},
  month = sep,
  pages = {1373–1375}
}

@article{qm9,
  title = {Quantum chemistry structures and properties of 134 kilo molecules},
  volume = {1},
  ISSN = {2052-4463},
  url = {http://dx.doi.org/10.1038/sdata.2014.22},
  DOI = {10.1038/sdata.2014.22},
  number = {1},
  journal = {Scientific Data},
  publisher = {Springer Science and Business Media LLC},
  author = {Ramakrishnan,  Raghunathan and Dral,  Pavlo O. and Rupp,  Matthias and von Lilienfeld,  O. Anatole},
  year = {2014},
  month = aug 
}

@article{mp_2013,
  title = {Commentary: The Materials Project: A materials genome approach to accelerating materials innovation},
  volume = {1},
  ISSN = {2166-532X},
  url = {http://dx.doi.org/10.1063/1.4812323},
  DOI = {10.1063/1.4812323},
  number = {1},
  journal = {APL Materials},
  publisher = {AIP Publishing},
  author = {Jain,  Anubhav and Ong,  Shyue Ping and Hautier,  Geoffroy and Chen,  Wei and Richards,  William Davidson and Dacek,  Stephen and Cholia,  Shreyas and Gunter,  Dan and Skinner,  David and Ceder,  Gerbrand and Persson,  Kristin A.},
  year = {2013},
  month = jul 
}

@article{oc20,
  title = {Open Catalyst 2020 (OC20) Dataset and Community Challenges},
  volume = {11},
  ISSN = {2155-5435},
  url = {http://dx.doi.org/10.1021/acscatal.0c04525},
  DOI = {10.1021/acscatal.0c04525},
  number = {10},
  journal = {ACS Catalysis},
  publisher = {American Chemical Society (ACS)},
  author = {Chanussot,  Lowik and Das,  Abhishek and Goyal,  Siddharth and Lavril,  Thibaut and Shuaibi,  Muhammed and Riviere,  Morgane and Tran,  Kevin and Heras-Domingo,  Javier and Ho,  Caleb and Hu,  Weihua and Palizhati,  Aini and Sriram,  Anuroop and Wood,  Brandon and Yoon,  Junwoong and Parikh,  Devi and Zitnick,  C. Lawrence and Ulissi,  Zachary},
  year = {2021},
  month = may,
  pages = {6059–6072}
}

@article{oc22,
  title = {The Open Catalyst 2022 (OC22) Dataset and Challenges for Oxide Electrocatalysts},
  volume = {13},
  ISSN = {2155-5435},
  url = {http://dx.doi.org/10.1021/acscatal.2c05426},
  DOI = {10.1021/acscatal.2c05426},
  number = {5},
  journal = {ACS Catalysis},
  publisher = {American Chemical Society (ACS)},
  author = {Tran,  Richard and Lan,  Janice and Shuaibi,  Muhammed and Wood,  Brandon M. and Goyal,  Siddharth and Das,  Abhishek and Heras-Domingo,  Javier and Kolluru,  Adeesh and Rizvi,  Ammar and Shoghi,  Nima and Sriram,  Anuroop and Therrien,  Félix and Abed,  Jehad and Voznyy,  Oleksandr and Sargent,  Edward H. and Ulissi,  Zachary and Zitnick,  C. Lawrence},
  year = {2023},
  month = feb,
  pages = {3066–3084}
}

@inproceedings{schnet_2017,
  title = {{{SchNet}}: {{A}} Continuous-Filter Convolutional Neural Network for Modeling Quantum Interactions},
  shorttitle = {{{SchNet}}},
  booktitle = {Advances in {{Neural Information Processing Systems}}},
  author = {Sch{\"u}tt, Kristof and Kindermans, Pieter-Jan and Felix, Huziel Enoc Sauceda and Chmiela, Stefan and Tkatchenko, Alexandre and M{\"u}ller, Klaus-Robert},
  year = {2017},
  pages = {992--1002}
}

@article{schnet_2018,
  title = {{{SchNet}} -- {{A}} Deep Learning Architecture for Molecules and Materials},
  author = {K. T. Sch{\"u}tt and H. E. Sauceda and {P. -J. Kindermans} and A. Tkatchenko and {K. -R. M{\"u}ller}},
  year = {2018},
  journal = {The Journal of Chemical Physics},
  volume = {148},
  number = {24},
  pages = {241722},
  doi = {10.1063/1.5019779}
}

@inproceedings{dimenet_2020,
  title = {Directional Message Passing for Molecular Graphs},
  booktitle = {8th International Conference on Learning Representations, {{ICLR}} 2020, Addis Ababa, Ethiopia, April 26-30, 2020},
  author = {Klicpera, Johannes and Gro{\ss}, Janek and G{\"u}nnemann, Stephan},
  year = {2020},
  publisher = {OpenReview.net},
  bibsource = {dblp computer science bibliography, https://dblp.org}
}

@article{nequip_2022,
  title = {E(3)-equivariant graph neural networks for data-efficient and accurate interatomic potentials},
  volume = {13},
  ISSN = {2041-1723},
  url = {http://dx.doi.org/10.1038/s41467-022-29939-5},
  DOI = {10.1038/s41467-022-29939-5},
  number = {1},
  journal = {Nature Communications},
  publisher = {Springer Science and Business Media LLC},
  author = {Batzner,  Simon and Musaelian,  Albert and Sun,  Lixin and Geiger,  Mario and Mailoa,  Jonathan P. and Kornbluth,  Mordechai and Molinari,  Nicola and Smidt,  Tess E. and Kozinsky,  Boris},
  year = {2022},
  month = may 
}

@inproceedings{mace_2022,
  title={{MACE}: Higher Order Equivariant Message Passing Neural Networks for Fast and Accurate Force Fields},
  author={Ilyes Batatia and David Peter Kovacs and Gregor N. C. Simm and Christoph Ortner and Gabor Csanyi},
  booktitle={Advances in Neural Information Processing Systems},
  editor={Alice H. Oh and Alekh Agarwal and Danielle Belgrave and Kyunghyun Cho},
  year={2022},
  url={https://openreview.net/forum?id=YPpSngE-ZU}
}

@misc{uma,
      title={UMA: A Family of Universal Models for Atoms}, 
      author={Brandon M. Wood and Misko Dzamba and Xiang Fu and Meng Gao and Muhammed Shuaibi and Luis Barroso-Luque and Kareem Abdelmaqsoud and Vahe Gharakhanyan and John R. Kitchin and Daniel S. Levine and Kyle Michel and Anuroop Sriram and Taco Cohen and Abhishek Das and Ammar Rizvi and Sushree Jagriti Sahoo and Zachary W. Ulissi and C. Lawrence Zitnick},
      year={2025},
      eprint={2506.23971},
      archivePrefix={arXiv},
      primaryClass={cs.LG},
      url={https://arxiv.org/abs/2506.23971}, 
}

@article{m3gnet,
  title = {A universal graph deep learning interatomic potential for the periodic table},
  volume = {2},
  ISSN = {2662-8457},
  url = {http://dx.doi.org/10.1038/s43588-022-00349-3},
  DOI = {10.1038/s43588-022-00349-3},
  number = {11},
  journal = {Nature Computational Science},
  publisher = {Springer Science and Business Media LLC},
  author = {Chen,  Chi and Ong,  Shyue Ping},
  year = {2022},
  month = nov,
  pages = {718–728}
}

@misc{mace_mp,
      title={A foundation model for atomistic materials chemistry}, 
      author={Ilyes Batatia and Philipp Benner and Yuan Chiang and Alin M. Elena and Dávid P. Kovács and Janosh Riebesell and Xavier R. Advincula and Mark Asta and Matthew Avaylon and William J. Baldwin and Fabian Berger and Noam Bernstein and Arghya Bhowmik and Filippo Bigi and Samuel M. Blau and Vlad Cărare and Michele Ceriotti and Sanggyu Chong and James P. Darby and Sandip De and Flaviano Della Pia and Volker L. Deringer and Rokas Elijošius and Zakariya El-Machachi and Fabio Falcioni and Edvin Fako and Andrea C. Ferrari and John L. A. Gardner and Mikolaj J. Gawkowski and Annalena Genreith-Schriever and Janine George and Rhys E. A. Goodall and Jonas Grandel and Clare P. Grey and Petr Grigorev and Shuang Han and Will Handley and Hendrik H. Heenen and Kersti Hermansson and Christian Holm and Cheuk Hin Ho and Stephan Hofmann and Jad Jaafar and Konstantin S. Jakob and Hyunwook Jung and Venkat Kapil and Aaron D. Kaplan and Nima Karimitari and James R. Kermode and Panagiotis Kourtis and Namu Kroupa and Jolla Kullgren and Matthew C. Kuner and Domantas Kuryla and Guoda Liepuoniute and Chen Lin and Johannes T. Margraf and Ioan-Bogdan Magdău and Angelos Michaelides and J. Harry Moore and Aakash A. Naik and Samuel P. Niblett and Sam Walton Norwood and Niamh O'Neill and Christoph Ortner and Kristin A. Persson and Karsten Reuter and Andrew S. Rosen and Louise A. M. Rosset and Lars L. Schaaf and Christoph Schran and Benjamin X. Shi and Eric Sivonxay and Tamás K. Stenczel and Viktor Svahn and Christopher Sutton and Thomas D. Swinburne and Jules Tilly and Cas van der Oord and Santiago Vargas and Eszter Varga-Umbrich and Tejs Vegge and Martin Vondrák and Yangshuai Wang and William C. Witt and Thomas Wolf and Fabian Zills and Gábor Csányi},
      year={2025},
      eprint={2401.00096},
      archivePrefix={arXiv},
      primaryClass={physics.chem-ph},
      url={https://arxiv.org/abs/2401.00096}, 
}

@article{rgd1,
  title = {Comprehensive exploration of graphically defined reaction spaces},
  volume = {10},
  ISSN = {2052-4463},
  url = {http://dx.doi.org/10.1038/s41597-023-02043-z},
  DOI = {10.1038/s41597-023-02043-z},
  number = {1},
  journal = {Scientific Data},
  publisher = {Springer Science and Business Media LLC},
  author = {Zhao,  Qiyuan and Vaddadi,  Sai Mahit and Woulfe,  Michael and Ogunfowora,  Lawal A. and Garimella,  Sanjay S. and Isayev,  Olexandr and Savoie,  Brett M.},
  year = {2023},
  month = mar 
}

@article{oc20_neb,
  title = {CatTSunami: Accelerating Transition State Energy Calculations with Pretrained Graph Neural Networks},
  volume = {15},
  ISSN = {2155-5435},
  url = {http://dx.doi.org/10.1021/acscatal.4c04272},
  DOI = {10.1021/acscatal.4c04272},
  number = {7},
  journal = {ACS Catalysis},
  publisher = {American Chemical Society (ACS)},
  author = {Wander,  Brook and Shuaibi,  Muhammed and Kitchin,  John R. and Ulissi,  Zachary W. and Zitnick,  C. Lawrence},
  year = {2025},
  month = mar,
  pages = {5283–5294}
}

@inproceedings{chemvlm,
author = {Li, Junxian and Zhang, Di and Wang, Xunzhi and Hao, Zeying and Lei, Jingdi and Tan, Qian and Zhou, Cai and Liu, Wei and Yang, Yaotian and Xiong, Xinrui and Wang, Weiyun and Chen, Zhe and Wang, Wenhai and Li, Wei and Su, Mao and Zhang, Shufei and Ouyang, Wanli and Li, Yuqiang and Zhou, Dongzhan},
title = {ChemVLM: exploring the power of multimodal large language models in chemistry area},
year = {2025},
isbn = {978-1-57735-897-8},
publisher = {AAAI Press},
url = {https://doi.org/10.1609/aaai.v39i1.32020},
doi = {10.1609/aaai.v39i1.32020},
booktitle = {Proceedings of the Thirty-Ninth AAAI Conference on Artificial Intelligence and Thirty-Seventh Conference on Innovative Applications of Artificial Intelligence and Fifteenth Symposium on Educational Advances in Artificial Intelligence},
articleno = {47},
numpages = {9},
series = {AAAI'25/IAAI'25/EAAI'25}
}

@inproceedings{crystal_llm,
  author       = {Nate Gruver and
                  Anuroop Sriram and
                  Andrea Madotto and
                  Andrew Gordon Wilson and
                  C. Lawrence Zitnick and
                  Zachary W. Ulissi},
  title        = {Fine-Tuned Language Models Generate Stable Inorganic Materials as
                  Text},
  booktitle    = {The Twelfth International Conference on Learning Representations,
                  {ICLR} 2024, Vienna, Austria, May 7-11, 2024},
  publisher    = {OpenReview.net},
  year         = {2024},
  url          = {https://openreview.net/forum?id=vN9fpfqoP1},
  bibsource    = {dblp computer science bibliography, https://dblp.org}
}

@misc{chem3dllm,
      title={Chem3DLLM: 3D Multimodal Large Language Models for Chemistry}, 
      author={Lei Jiang and Shuzhou Sun and Biqing Qi and Yuchen Fu and Xiaohua Xu and Yuqiang Li and Dongzhan Zhou and Tianfan Fu},
      year={2025},
      eprint={2508.10696},
      archivePrefix={arXiv},
      primaryClass={cs.CE},
      url={https://arxiv.org/abs/2508.10696}, 
}

@misc{chemllm,
      title={ChemLLM: A Chemical Large Language Model}, 
      author={Di Zhang and Wei Liu and Qian Tan and Jingdan Chen and Hang Yan and Yuliang Yan and Jiatong Li and Weiran Huang and Xiangyu Yue and Wanli Ouyang and Dongzhan Zhou and Shufei Zhang and Mao Su and Han-Sen Zhong and Yuqiang Li},
      year={2024},
      eprint={2402.06852},
      archivePrefix={arXiv},
      primaryClass={cs.AI},
      url={https://arxiv.org/abs/2402.06852}, 
}

@phdthesis{uspto, title={Extraction of chemical structures and reactions from the literature}, url={https://www.repository.cam.ac.uk/handle/1810/244727}, DOI={10.17863/CAM.16293}, school={Apollo - University of Cambridge Repository}, author={Lowe, Daniel Mark}, year={2012} }

@article{tran1x,
  title = {Transition1x - a dataset for building generalizable reactive machine learning potentials},
  volume = {9},
  ISSN = {2052-4463},
  url = {http://dx.doi.org/10.1038/s41597-022-01870-w},
  DOI = {10.1038/s41597-022-01870-w},
  number = {1},
  journal = {Scientific Data},
  publisher = {Springer Science and Business Media LLC},
  author = {Schreiner,  Mathias and Bhowmik,  Arghya and Vegge,  Tejs and Busk,  Jonas and Winther,  Ole},
  year = {2022},
  month = dec 
}

@inproceedings{cm_2012,
    author = {Montavon, Gr{\'e}goire and Hansen, Katja and Fazli, Siamac and Rupp, Matthias and Biegler, Franziska and Ziehe, Andreas and Tkatchenko, Alexandre and {von Lilienfeld}, Anatole and M{\"u}ller, Klaus-Robert},
    editor = "Bartlett, Peter L. and Pereira, Fernando C.N. and Burges, Christopher J.C. and Bottou, L{\'e}on and Weinberger, Kilian Q.",
    title = "Learning Invariant Representations of Molecules for Atomization Energy Prediction",
    booktitle = "Advances in {{Neural Information Processing Systems}} 25: 26th {{Annual Conference}} on {{Neural Information Processing Systems}} 2012. {{Proceedings}} of a Meeting Held {{December}} 3-6, 2012, {{Lake Tahoe}}, {{Nevada}}, {{United States}}",
    year = "2012",
    pages = "449--457"
}

@article{e3fp,
  title = {A Simple Representation of Three-Dimensional Molecular Structure},
  volume = {60},
  ISSN = {1520-4804},
  url = {http://dx.doi.org/10.1021/acs.jmedchem.7b00696},
  DOI = {10.1021/acs.jmedchem.7b00696},
  number = {17},
  journal = {Journal of Medicinal Chemistry},
  publisher = {American Chemical Society (ACS)},
  author = {Axen,  Seth D. and Huang,  Xi-Ping and Cáceres,  Elena L. and Gendelev,  Leo and Roth,  Bryan L. and Keiser,  Michael J.},
  year = {2017},
  month = aug,
  pages = {7393–7409}
}

@article{uspto_500k,
  title = {Improving Few- and Zero-Shot Reaction Template Prediction Using Modern Hopfield Networks},
  volume = {62},
  ISSN = {1549-960X},
  url = {http://dx.doi.org/10.1021/acs.jcim.1c01065},
  DOI = {10.1021/acs.jcim.1c01065},
  number = {9},
  journal = {Journal of Chemical Information and Modeling},
  publisher = {American Chemical Society (ACS)},
  author = {Seidl,  Philipp and Renz,  Philipp and Dyubankova,  Natalia and Neves,  Paulo and Verhoeven,  Jonas and Wegner,  J\"{o}rg K. and Segler,  Marwin and Hochreiter,  Sepp and Klambauer,  G\"{u}nter},
  year = {2022},
  month = jan,
  pages = {2111–2120}
}

@article{scaffold_split,
  title = {The Properties of Known Drugs. 1. Molecular Frameworks},
  volume = {39},
  ISSN = {1520-4804},
  url = {http://dx.doi.org/10.1021/jm9602928},
  DOI = {10.1021/jm9602928},
  number = {15},
  journal = {Journal of Medicinal Chemistry},
  publisher = {American Chemical Society (ACS)},
  author = {Bemis,  Guy W. and Murcko,  Mark A.},
  year = {1996},
  month = jan,
  pages = {2887–2893}
}

@misc{qwen,
      title={Qwen3 Technical Report}, 
      author={An Yang and Anfeng Li and Baosong Yang and Beichen Zhang and Binyuan Hui and Bo Zheng and Bowen Yu and Chang Gao and Chengen Huang and Chenxu Lv and Chujie Zheng and Dayiheng Liu and Fan Zhou and Fei Huang and Feng Hu and Hao Ge and Haoran Wei and Huan Lin and Jialong Tang and Jian Yang and Jianhong Tu and Jianwei Zhang and Jianxin Yang and Jiaxi Yang and Jing Zhou and Jingren Zhou and Junyang Lin and Kai Dang and Keqin Bao and Kexin Yang and Le Yu and Lianghao Deng and Mei Li and Mingfeng Xue and Mingze Li and Pei Zhang and Peng Wang and Qin Zhu and Rui Men and Ruize Gao and Shixuan Liu and Shuang Luo and Tianhao Li and Tianyi Tang and Wenbiao Yin and Xingzhang Ren and Xinyu Wang and Xinyu Zhang and Xuancheng Ren and Yang Fan and Yang Su and Yichang Zhang and Yinger Zhang and Yu Wan and Yuqiong Liu and Zekun Wang and Zeyu Cui and Zhenru Zhang and Zhipeng Zhou and Zihan Qiu},
      year={2025},
      eprint={2505.09388},
      archivePrefix={arXiv},
      primaryClass={cs.CL},
      url={https://arxiv.org/abs/2505.09388}, 
}

@article{Morgan1965,
  title = {The Generation of a Unique Machine Description for Chemical Structures-A Technique Developed at Chemical Abstracts Service.},
  volume = {5},
  ISSN = {1541-5732},
  url = {http://dx.doi.org/10.1021/c160017a018},
  DOI = {10.1021/c160017a018},
  number = {2},
  journal = {Journal of Chemical Documentation},
  publisher = {American Chemical Society (ACS)},
  author = {Morgan,  H. L.},
  year = {1965},
  month = may,
  pages = {107–113}
}

@article{achiam2023gpt,
  title={Gpt-4 technical report},
  author={Achiam, Josh and Adler, Steven and Agarwal, Sandhini and Ahmad, Lama and Akkaya, Ilge and Aleman, Florencia Leoni and Almeida, Diogo and Altenschmidt, Janko and Altman, Sam and Anadkat, Shyamal and others},
  journal={arXiv preprint arXiv:2303.08774},
  year={2023}
}

@article{chen2024weak,
  title={Weak-eval-strong: Evaluating and eliciting lateral thinking of LLMs with situation puzzles},
  author={Chen, Qi and Zhang, Bowen and Wang, Gang and Wu, Qi},
  journal={Advances in Neural Information Processing Systems},
  volume={37},
  pages={79642--79665},
  year={2024}
}

@inproceedings{yang2025separation,
  title={Separation of powers: On segregating knowledge from observation in LLM-enabled knowledge-based visual question answering},
  author={Yang, Zhen and Tao, Zhuo and Chen, Qi and Li, Liang and Qi, Yuankai and Van Den Hengel, Anton and Huang, Qingming},
  booktitle={Proceedings of the Computer Vision and Pattern Recognition Conference},
  pages={24753--24762},
  year={2025}
}

@inproceedings{ren2025efficiently,
  title={Efficiently Selecting Response Generation Strategies for Synthetic Data Construction by Self-Aligned Perplexity},
  author={Ren, Xuan and Chen, Qi and Liu, Lingqiao},
  booktitle={Findings of the Association for Computational Linguistics: EMNLP 2025},
  pages={11584--11605},
  year={2025}
}
\bibliographystyle{icml2026}

\newpage
\appendix
\onecolumn
\section{Chem4DBench}
\label{appen:bench}
We provide further details in constructing the Chem4D benchmark

\subsection{Reaction Product Prediction}
Chemical reactions represent a fundamentally 4D scenario, where atoms evolve from initial reactant positions through high-energy transition states to final product geometries. To decouple the understanding of reaction mechanisms from simple 2D pattern matching, we constructed the Reaction Product Prediction task using two complementary quantum-chemical datasets: Transition1x \citep{tran1x} and RGD1 \citep{rgd1}. For both datasets, we constructed the task by providing the LLM with two 3D frames—the initial reactant geometries and the TS geometries—and asking it to predict the product SMILES, reaction barrier, and reaction enthalpy. To evaluate the model's performance in various OOD cases, we constructed four distinct evaluation splits: an in-distribution (ID) split and three out-of-distribution (OOD) splits—OOD-Reactants, OOD-Products, and OOD-Both. These OOD splits are determined via scaffold partitioning, ensuring that the evaluation probes the model's ability to extrapolate to molecular frameworks that are structurally distinct from those seen during training.

\begin{figure*}[h]
    \centering
    \includegraphics[width=1.0\textwidth]{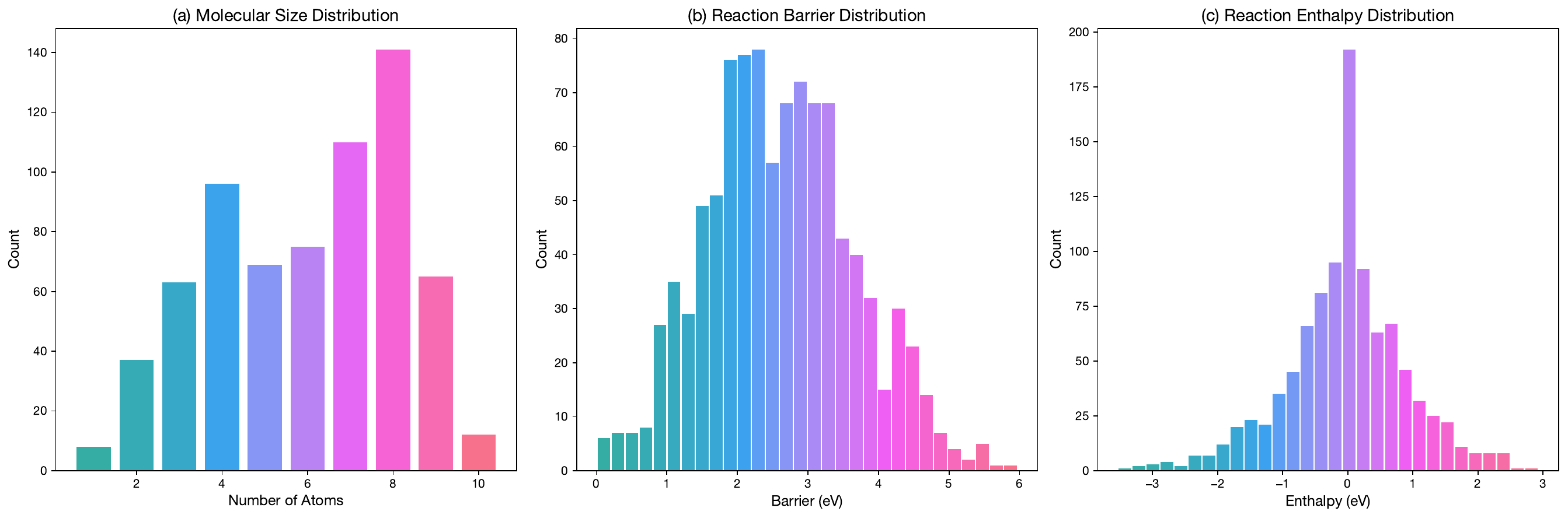}
    \caption{Statistical distribution of Reaction Product Prediction in the Chem4D benchmark. (a) The distribution of the number of atoms. (b) The distribution of the reaction barrier (eV). (c) The distribution of the reaction enthalpy (eV).}
    \label{fig:reaction_stats}
\end{figure*}

\subsection{Catalytic Reaction}
Heterogeneous catalysis involves complex spatio-temporal processes where reactants adsorb onto extended surfaces, undergo bond rearrangements mediated by the catalyst, and subsequently desorb as products. Compared to gas-phase or isolated molecular reactions, catalytic systems introduce additional challenges, including PBC (PBC), surface--adsorbate interactions, and collective atomic rearrangements of the catalyst lattice. To capture these characteristics, we construct the Catalytic Reactions category, which focuses on surface-mediated reaction dynamics in periodic environments.

We build this subset based on the OC20-NEB dataset \citep{oc20_neb}, an extension of Open Catalyst 2020 \citep{oc20} that provides nudged elastic band (NEB) trajectories for surface reactions. OC20-NEB explicitly resolves transition pathways for three key classes of elementary surface reactions---transfer, dissociation, and desorption---by tracing the atomic-scale evolution from reactant states, through transition-state (TS) regions, to product configurations.

To overcome the limited scale of OC20-NEB ($\sim$ 700 samples), we further augment this category with approximately 6{,}000 additional surface-reaction trajectories. We start by sampling catalyst--adsorbate reactant states from OC20 \citep{oc20}. For each sampled structure, we randomly assign a reaction class (transfer, dissociation, or desorption) and re-simulate the corresponding pathway with NEB optimization using the pretrained UMA model as the underlying surrogate potential. We record energies along the reaction coordinate to identify the TS frame, reaction barrier ($\Delta E^\ddagger$), and reaction enthalpy ($\Delta H$). This augmentation yields a larger and more diverse set of kinetically meaningful surface reaction pathways, substantially strengthening evaluation of 4D 4D understanding under PBC.

\begin{figure*}[h]
    \centering
    \includegraphics[width=1.0\textwidth]{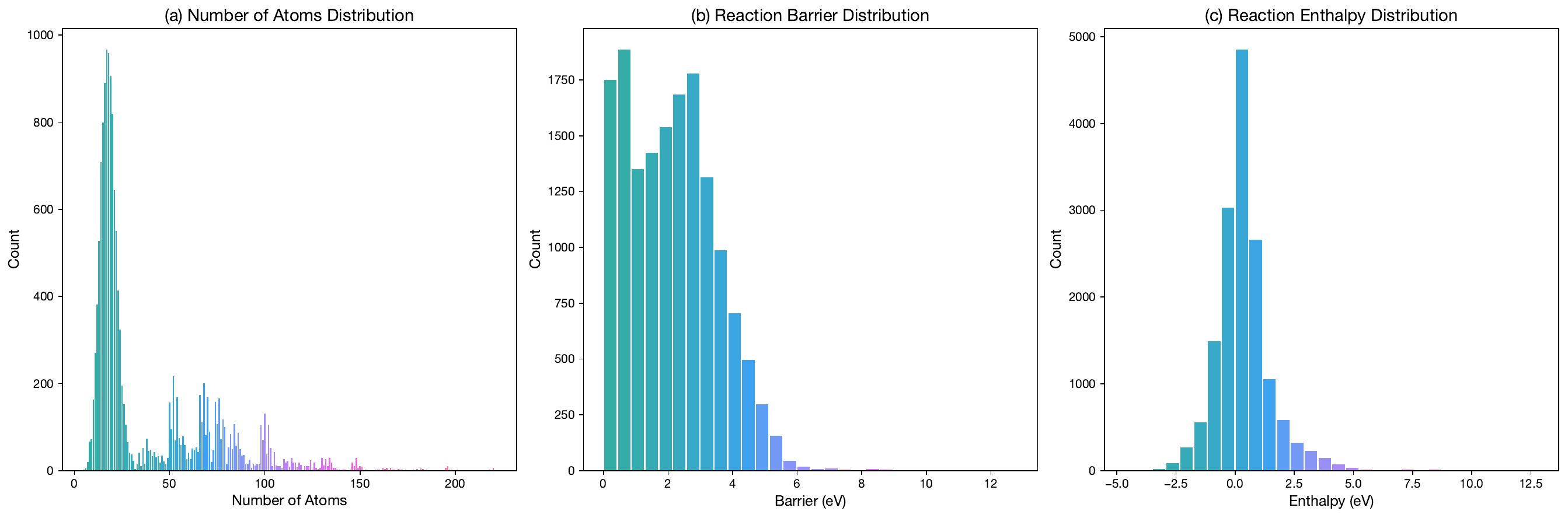}
    \caption{Statistical distribution of Catalytic Reaction Understanding in the Chem4D benchmark. (a) The distribution of the number of atoms. (b) The distribution of the reaction barrier (eV). (c) The distribution of the reaction enthalpy (eV).}
    \label{fig:reaction_stats}
\end{figure*}

\section{Evaluation Protocol}
\label{app:eval}

Evaluation is performed separately for each Chem4D category, with metrics designed to reflect both linguistic correctness and underlying chemical and physical validity, as summarized in Tables~\ref{tab:t1x}, \ref{tab:oc20_results}.

\paragraph{Reaction Product Prediction.}
For gas-phase reaction product prediction, we measure product SMILES accuracy using EXACT match and Levenshtein distance, capturing strict correctness and string-level edit error.
Second, we compute structural similarity between predicted and ground-truth products using fingerprint metrics: MACCS, RDK, and Morgan.
Third, we report VALIDITY, defined as the fraction of predicted SMILES that are chemically valid (i.e., can be parsed into a molecular graph).
Fourth, we evaluate intrinsic physical targets---reaction barrier ($\Delta E^\ddagger$) and reaction enthalpy ($\Delta H$)---using mean absolute error (MAE) in eV.
Finally, we report sequence-level BLEU to quantify overall similarity between predicted and reference SMILES.
Performance is reported on both in-domain (ID) and out-of-distribution (OOD) splits to evaluate extrapolative generalization.

\paragraph{Catalytic Reactions.}
For heterogeneous catalysis under PBC, evaluation focuses on joint reasoning over surface structure, reaction type, and energetics. As shown in Table~\ref{tab:oc20_results}, we report reaction type accuracy (transfer, dissociation, desorption), Adsorbate EXACT and Product EXACT match rates, together with their corresponding Morgan similarities. Reaction barriers and enthalpies are again evaluated using MAE. These metrics collectively assess whether the model correctly identifies the catalytic event, predicts physically correct surface-bound structures, and captures the associated energetic landscape.



\section{Details of baselines}
\label{appen:model}
\subsection{Chem3DLLM}
We consider Chem3DLLM \citep{chem3dllm} baseline that converts molecular geometries into structured natural-language descriptions, enabling a standard large language model to consume 3D information without any dedicated geometric encoder. This design is intentionally lightweight and contrasts with multimodal architectures that rely on learned 3D embeddings or equivariant networks.

\textbf{Difference from SDF-to-text in Chem3DLLM.}
Unlike the SDF-to-text encoding used in Chem3DLLM, which performs a includs exact bond orders and coordinates via compressed tokenization, the 3D text-based baseline adopted here is descriptive rather than generative. It does not aim to reconstruct the original structure from text, nor does it preserve complete bonding information. Instead, it provides a human-readable geometric summary—atomic identities, Cartesian coordinates, and a bounded set of local interatomic distances—sufficient for downstream reasoning tasks such as adsorption understanding and reaction description. As a result, this baseline trades reversibility and completeness for simplicity and prompt efficiency.

\textbf{Handling periodic boundary conditions.}
For periodic systems, which are common in heterogeneous catalysis, the text-based baseline explicitly exposes lattice information rather than attempting to encode periodicity implicitly. When PBC are present, the $3\times3$ lattice matrix is printed as three basis vectors labeled by the $x$, $y$, and $z$ directions. Atomic positions are reported in Cartesian coordinates within the reference cell.

Neighbor distances under PBC are computed using a radius-based search that accounts for periodic images. In practice, the unit cell is tiled sufficiently to ensure that all neighbors within the cutoff radius are captured. For slab geometries, periodicity is applied only in the surface plane, while the surface-normal direction remains non-periodic. After candidate neighbors are generated from periodic images, a strict cap on the maximum number of neighbors per atom is enforced to maintain a bounded and consistent textual representation.

\subsection{4D Text-based}
We modify Chem3DLLM to 4D Text-based by representing a molecular trajectory as an ordered sequence of frames, where $\mathcal{X}_0$ denotes the initial state and $\mathcal{X}_t$ denotes the configuration at time step $t$. To control prompt length and ensure scalability to long trajectories, we adopt a compact text-based encoding that combines numeric truncation with sparse temporal updates.

\textbf{Initial State Encoding ($\mathcal{X}_0$).}
The first frame provides a complete description of the system. For each atom, we record its chemical symbol and absolute Cartesian coordinates $(x, y, z)$, rounded to two decimal places. For periodic systems, the $3\times3$ lattice matrix is explicitly included as three basis vectors.  

To expose local geometric context, we construct a radius-based neighborhood graph with a cutoff of $2.5\,\text{\AA}$. To bound the textual length, we cap the maximum number of neighbors per atom at $k=4$, retaining only the nearest neighbors. For systems under PBC, distances are computed using the minimum image convention. The resulting neighborhood information is serialized as tuples of the form \texttt{(src\_idx, tgt\_idx, distance)}, where all distances are also rounded to two decimal places.

\textbf{Sparse Differential Encoding ($\mathcal{X}_{t>0}$).}
For subsequent frames, we apply a sparse, event-driven encoding strategy. Instead of re-listing the full structure, we include only atoms whose displacement exceeds a predefined threshold. Specifically, an atom $i$ is reported at frame $t$ only if
\[
\|\mathbf{r}_i^{(t)} - \mathbf{r}_i^{(t-1)}\| > \delta,\quad \delta = 0.1\,\text{\AA}.
\]

For each such atom, we report: 1st, Geometry updates: the updated Cartesian coordinates, rounded to two decimal places. 2nd, Topology updates: newly computed neighbor distances, but only for edges incident to atoms that satisfy the displacement criterion.

Atoms that do not exceed the displacement threshold are omitted entirely from the frame description. This design significantly reduces token consumption for rigid or inactive regions, while preserving high-resolution information around dynamically evolving reaction centers.

The final input to the language model is formed by concatenating the textual descriptions of $\mathcal{X}_0$ and all subsequent sparse updates $\{\mathcal{X}_t\}_{t>0}$. This representation allows the LLM to reason over temporal evolution while remaining within a practical context window.

Table~\ref{tab:chem3d_vs_4dtext} provides the examples of Chem3DLLM and 4D Text-based methods.

\begin{table*}[t]
\caption{Example inputs for Chem3D-style (3D) and 4D-text-based models. Both settings aim to predict the product SMILES, reaction barrier, and reaction enthalpy, but differ in how geometric information is provided.}
\label{tab:chem3d_vs_4dtext}
\centering
\scriptsize
\begin{tabular}{p{7.2cm} p{7.2cm}}
\toprule
\textbf{Chem3DLLLM-style (3D Input)} & \textbf{4D-Text-based (Trajectory Input)} \\
\midrule
\begin{verbatim}
Given the reactant and transition state geometries, 
predict the product SMILES, reaction barrier and 
reaction enthalpy. The geometry is as follows:
The system is non-periodic. The atom and its 
position are as follows:
C1 -1.6169 0.5542 0.9918
C2 -2.2864 -0.7707 1.3265
C3 -1.4256 0.7574 -0.4801
...
H21 1.0702 -0.4440 -2.0669
The distance between atoms is as follows:
atom1_index atom2_index distance(A)
1 7 1.0930
1 6 1.0964
...
21 19 2.3363
\end{verbatim}
&
\begin{verbatim}
Given the reactant and transition state geometries, 
predict the product SMILES, reaction barrier and 
reaction enthalpy. The geometry is as follows:
Frame 1:
The system is non-periodic. The atom and its 
position are as follows:
C1 -1.62 0.55 0.99
C2 -2.29 -0.77 1.33
...
The distance between atoms is as follows:
atom1_index atom2_index distance(A)
1 7 1.09
1 6 1.10
...
21 19 2.34

Frame 2:
The atom with big movements are listed as follows:
C1 0.77 1.22 -0.69
C2 2.26 1.20 -0.37
...
H21 -1.48 -1.01 1.37
New edge distances:
atom1_index atom2_index distance(A)
1 7 1.09
1 6 1.09
...
21 19 2.25
\end{verbatim}
\\
\bottomrule
\end{tabular}
\end{table*}

\paragraph{4D-MolT5.}
We further develop \textbf{4D-MolT5} as a temporal extension of 3D-MolT5, designed to interpret molecular dynamics and reaction pathways by lifting static spatial encoding into the temporal domain. 
In 3D-MolT5, a 1D SELFIES sequence is aligned with a single static 3D molecular conformation, which limits its ability to represent conformational ensembles, transition states, or trajectory-dependent phenomena. 
In contrast, 4D-MolT5 processes an entire molecular trajectory.

Formally, a 4D trajectory is represented as a sequence of $T$ frames,
\[
\mathcal{D} = \{D^{(1)}, D^{(2)}, \dots, D^{(T)}\},
\]
where each $D^{(t)}$ corresponds to the 3D token sequence of the molecule at time step $t$. 
To capture spatiotemporal dependencies without modifying the underlying T5 encoder architecture, we flatten the temporal dimension into the token sequence dimension. 
This results in an extended input sequence of length $T \times N$, where $N$ is the number of atoms.

To preserve atom-level semantic alignment across time, the corresponding 1D SELFIES-based semantic tokens are broadcasted $T$ times and paired with each frame. 
This serialization strategy allows the global self-attention mechanism of T5 to jointly model intra-frame spatial geometry and inter-frame temporal evolution, effectively enabling the model to \emph{observe} molecular dynamics and reaction trajectories as a unified 4D sequence.

While this approach allows 4D-MolT5 to ingest time-resolved structural information using an unmodified language backbone, it relies on sequence flattening and lacks explicit mechanisms to preserve atomic identity or enforce physical inductive biases across frames, which can limit scalability and fine-grained causal reasoning for long trajectories or large systems.

\section{Output Example}
\begin{figure}[h]
    \centering
    \includegraphics[width=0.5\linewidth]{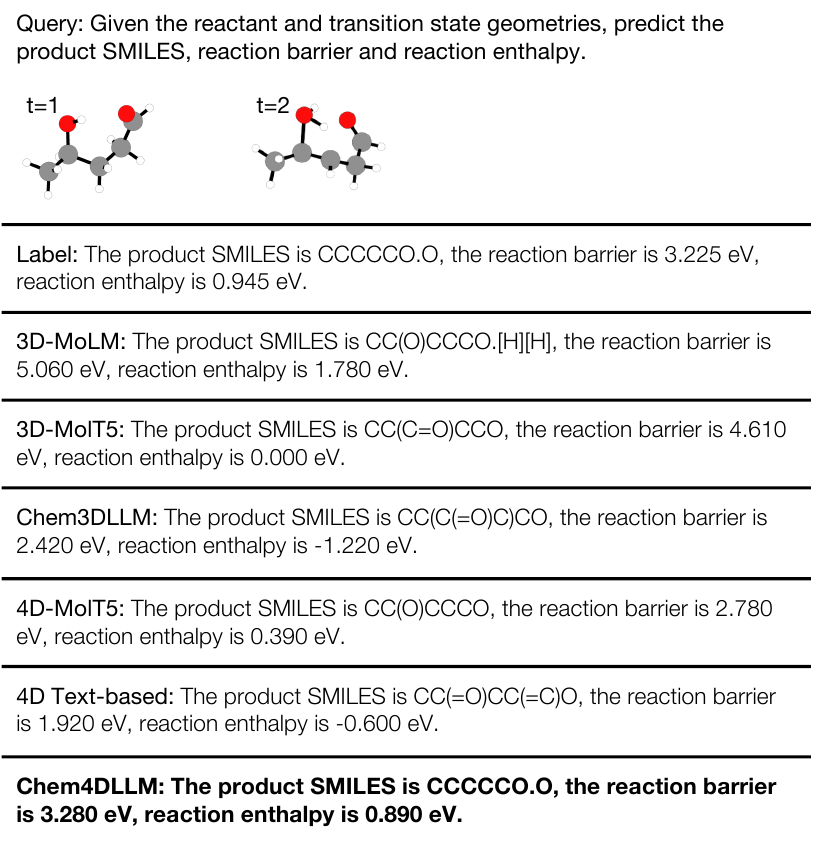}
    \caption{Sample of Reaction Product Prediction}
    \label{fig:sample_t1x}
\end{figure}

\begin{figure}[h]
    \centering
    \includegraphics[width=0.5\linewidth]{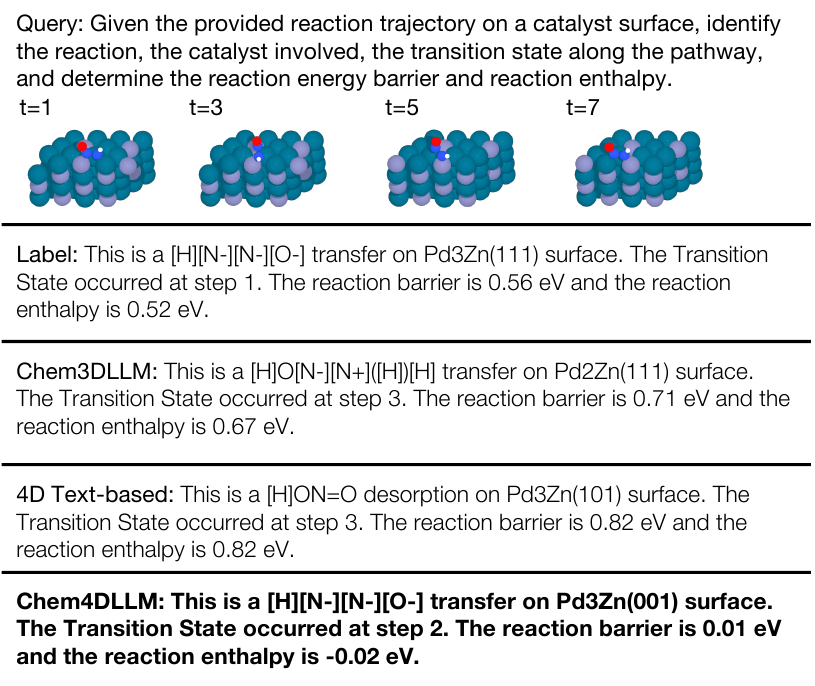}
    \caption{Sample of Catalytic Reaction Understanding}
    \label{fig:sample_oc20}
\end{figure}

\end{document}